%% file: _main_acm_survey.tex
\documentclass[prologue,table,usenames,dvipsnames,manuscript,screen]{acmart}

\AtBeginDocument{%
  }

\setcopyright{acmlicensed}
\copyrightyear{2018}
\acmYear{2018}
\acmDOI{XXXXXXX.XXXXXXX}
\acmISBN{978-1-4503-XXXX-X/2018/06}

\usepackage{tabularx}
\usepackage{booktabs}
\usepackage{array}
\usepackage{enumitem}
\input{sections_journal/_preamble}
\begin{document}

\title{Game Theory Meets Large Language Models: A Systematic Survey with Taxonomy and New Frontiers}


% \author{Authors}
% \email{Email}
% \affiliation{%
%   \institution{university}
%   \city{Beijing}
%   \country{China}
% }

\author{Haoran Sun}
\email{sunhaoran0301@stu.pku.edu.cn}
\affiliation{%
  \institution{CFCS, School of Computer Science, Peking University}
  \city{Beijing}
  \country{China}
}

\author{Yusen Wu}
\email{sarinice2025@stu.pku.edu.cn}
\affiliation{%
  \institution{CFCS, School of Computer Science, Peking University}
  \city{Beijing}
  \country{China}
}

\author{Peng Wang}
\email{wp@stu.jiangnan.edu.cn}
\affiliation{%
  \institution{School of Business, Jiangnan University}
  \city{Wuxi}
  \state{Jiangsu}
  \country{China}
}

\author{Wei Chen}
\email{weic@microsoft.com}
\affiliation{%
  \institution{Microsoft Research Asia}
  \city{Beijing}
  \country{China}
}

\author{Yukun Cheng}
\email{ykcheng@amss.ac.cn}
\authornote{Corresponding Authors.}
\affiliation{%
  \institution{School of Business, Jiangnan University}
  \city{Wuxi}
  \state{Jiangsu}
  \country{China}
}

\author{Xiaotie Deng}
\email{xiaotie@pku.edu.cn}
\authornotemark[1]
\affiliation{%
  \institution{CFCS, School of Computer Science, Peking University}
  \city{Beijing}
  \country{China}
}

\author{Xu Chu}
\email{chu\_xu@pku.edu.cn}
\authornotemark[1]
\affiliation{%
  \institution{CFCS, School of Computer Science, Peking University}
  \city{Beijing}
  \country{China}
}

\renewcommand{\shortauthors}{Sun et al.}

\input{sections_journal/0_abstract}

\begin{CCSXML}
<ccs2012>
   <concept>
       <concept_id>10002944.10011122.10002945</concept_id>
       <concept_desc>General and reference~Surveys and overviews</concept_desc>
       <concept_significance>500</concept_significance>
       </concept>
   <concept>
       <concept_id>10010147.10010178</concept_id>
       <concept_desc>Computing methodologies~Artificial intelligence</concept_desc>
       <concept_significance>500</concept_significance>
       </concept>
   <concept>
       <concept_id>10003752.10010070.10010099.10010100</concept_id>
       <concept_desc>Theory of computation~Algorithmic game theory</concept_desc>
       <concept_significance>300</concept_significance>
       </concept>
 </ccs2012>
\end{CCSXML}

\ccsdesc[500]{General and reference~Surveys and overviews}
\ccsdesc[500]{Computing methodologies~Artificial intelligence}
\ccsdesc[300]{Theory of computation~Algorithmic game theory}

\keywords{Large Language Models, Game Theory, Machine Learning}

\received{20 February 2007}
\received[revised]{12 March 2009}
\received[accepted]{5 June 2009}

\maketitle

\input{sections_journal/1_intro}

\input{sections_journal/2_game_for_llm_evaluation}
\input{sections_journal/3_game_for_llm_algorithms}

\input{sections_journal/4_game_models_for_llm}
\input{sections_journal/5_llm_for_game}
\input{sections_journal/6_future_directions}
\input{sections_journal/7_conclusion}

\bibliographystyle{unsrtnat}
\bibliography{reference}
% \appendix

% \nocite{*}
% \input{sections_journal/ref}
\end{document}

%% file: sections_journal/_preamble.tex
\usepackage{microtype}
\usepackage{amsfonts} % defines \Bbbk
\usepackage{bbm}      % now can safely define \Bbbk
\usepackage{amssymb}
\usepackage{subfig}
\usepackage{float}
\usepackage{multirow}
\usepackage{tikz}
\usepackage[disable]{todonotes}
\usepackage{tcolorbox}

\usetikzlibrary{fadings}
\usetikzlibrary{mindmap,trees}
\usetikzlibrary{arrows,automata,shapes,positioning,shadows,trees}
\usetikzlibrary{shapes.arrows}
\usetikzlibrary{calc,shapes, positioning}
\usetikzlibrary{decorations.text}
\usetikzlibrary{matrix,chains,positioning,decorations.pathreplacing,arrows}
\usetikzlibrary{bayesnet}
\usepackage[edges]{forest}
\usetikzlibrary{shadows.blur}
\usetikzlibrary{shapes.geometric}

\usepackage{booktabs}
\usepackage{makecell}
\usepackage{rotating}

\usepackage{pgfplots}
\usepackage{pgfplotstable}
\usepackage{pgfmath,pgffor}
\usepgfplotslibrary{colorbrewer}
\pgfplotsset{compat = 1.14, cycle list/Set1-8}
\usetikzlibrary{pgfplots.statistics, pgfplots.colorbrewer}
\pgfplotsset{compat=1.8}

\tikzstyle{edge}=[-latex',draw=black!90,shorten <=1pt,shorten >=1pt]
\tikzstyle{redge}=[latex'-,draw=black!90,shorten <=1pt,shorten >=1pt]
\tikzstyle{dedge}=[latex'-latex',draw=black!90,shorten <=1pt,shorten >=1pt]

\tikzstyle{block}=[draw, text width=5em,align=center,shape=rectangle, rounded corners, , align=center]
\tikzstyle{nobox}=[align=center]
\definecolor{emb}{RGB}{209,228,252}
\definecolor{hidden-blue}{RGB}{194,232,247}
\definecolor{hidden-orange}{RGB}{224,224,224}
\definecolor{hidden-yellow}{RGB}{242,244,193}
\definecolor{output-purple}{RGB}{219,203,231}
\definecolor{output-green}{RGB}{204,231,207}
\definecolor{hiddendraw}{RGB}{10,128,122}
\definecolor{lightgray}{gray}{0.95}

\usetikzlibrary{arrows.meta}    % 加载TikZ的箭头库，以使用 -latex 箭头样式

\definecolor{rootcolor}{RGB}{168, 108, 98}      % 定义根节点填充色 (陶土棕红)
\definecolor{roottext}{RGB}{255, 255, 255}      % 定义根节点文字颜色 (白色)
\definecolor{level1color}{RGB}{210, 170, 153}    % 定义第一层节点填充色 (裸粉色/干枯玫瑰)
\definecolor{level2color}{RGB}{244, 238, 224}    % 定义第二层节点填充色 (暖米灰色)
\definecolor{edgecolor}{RGB}{89, 74, 66}        % 定义边和节点边框的颜色 (深咖啡)

\makeatletter
\renewcommand\paragraph{\@startsection{paragraph}{4}{\z@}%
{1ex \@plus1ex \@minus.2ex}%
 {-1em}%
 {\normalfont\normalsize\bfseries}}
\makeatother

\usepackage{cleveref}

%% file: sections_journal/0_abstract.tex
\begin{abstract}

Game theory is a foundational framework for analyzing strategic interactions, and its intersection with large language models (LLMs) is a rapidly growing field. However, existing surveys mainly focus narrowly on using game theory to evaluate LLM behavior. This paper provides the first comprehensive survey of the bidirectional relationship between Game Theory and LLMs. We propose a novel taxonomy that categorizes the research in this intersection into four distinct perspectives: (1) evaluating LLMs in game-based scenarios; (2) improving LLMs using game-theoretic concepts for better interpretability and alignment; (3) modeling the competitive landscape of LLM development and its societal impact; and (4) leveraging LLMs to advance game models and to solve corresponding game theory problems. Furthermore, we identify key challenges and outline future research directions. By systematically investigating this interdisciplinary landscape, our survey highlights the mutual influence of game theory and LLMs, fostering progress at the intersection of these fields.
\end{abstract}

%% file: sections_journal/1_intro.tex
\section{Introduction}\label{sec:intro}

Game theory provides a mathematical framework for analyzing strategic interactions among rational agents, and it has evolved significantly since its seminal work~\cite{von2007theory}. 
Over the decades, it has established robust methodological foundations, including equilibrium analysis~\cite{nash1950equilibrium}, mechanism design~\cite{vickrey1961counterspeculation}, information design~\cite{bergemann2019information}, and social choice theory~\cite{sen1986social}. 
These concepts serve as essential analytical tools across diverse disciplines such as economics, political science, and computer science, offering insights into decision-making in competitive and cooperative environments. 
More recently, this field has also contributed to artificial intelligence~\cite{zhu2021survey,hazra-anjaria2022applications}, particularly in multi-agent systems and algorithmic game theory, as AI systems increasingly interact in complex ways.

The rapid advancement of large language models has revolutionized natural language processing and artificial intelligence~\cite{touvron2023llama,achiam2023gpt,team2023gemini,liu2024deepseek}. 
With their remarkable capabilities in language comprehension, generation, and reasoning, LLMs are increasingly integrated into various applications and have opened new avenues for research. 
This growing attention on LLMs has led to a surge of studies at the intersection of game theory and LLMs. 
Specifically, various work investigates \emph{how game-theoretic principles can enhance LLM evaluation and development, as well as how LLMs can contribute to advancing game theory itself}.

In this survey, we categorize these research efforts into four key directions:
\begin{itemize}
    \item \textbf{Evaluating LLMs in Game-based Playgrounds (Section~\ref{sec:game_for_llm_evaluation}):} This area focuses on constructing game-based benchmark environments, such as foundational matrix games~\cite{akata2023playing} and communication-based games like Avalon~\cite{wang2023avalon}, bargaining~\cite{deng2024bargaining}, and auctions~\cite{chen2023put}, to systematically evaluate the strategic reasoning capabilities of LLMs. 
    Researchers are also investigating how advanced techniques like prompt engineering~\cite{wang2023avalon, zhang2025k}, training~\cite{huang2024poker, fu2023improvinglanguagemodelnegotiation}, and tool-using~\cite{xia2024measuring, wu2024deciphering} influence LLM performance in these strategic contexts.
    \item \textbf{Improving LLMs with Game-theoretic Methods (Section~\ref{sec:game_for_llm_algorithm}):} This direction explores how concepts from cooperative and non-cooperative game theory, such as Shapley Value~\cite{enouen2023textgenshap}, social choice theory~\cite{ge2024axioms}, and max-min equilibria~\cite{NLHF}, can be utilized to design more efficient and theoretically sound algorithms for LLMs. Game-theoretic methods offer potential solutions for key LLM challenges, including model interpretability, general preference alignment, heterogeneity, and dynamic adaptation.
    \item \textbf{Characterizing LLM-related Events through Game Models (Section~\ref{sec:game_models}):} As LLMs are emerging technologies profoundly impacting human society and production, researchers are constructing game models to characterize events related to their development and deployment. This includes studies modeling the competition among multiple stakeholders in LLM development~\cite{laufer2024fine,duetting2024mechanism}, as well as those focusing on the societal impact of LLMs through game-theoretic lenses~\cite{yao2024human,taitler2024braess}.
    \item \textbf{Advancing Game Theory with LLMs (Section~\ref{sec:llm_for_game}):} Leveraging their superior natural language comprehension and generation capabilities, LLMs are being employed to advance classic game theory. Specifically, LLMs can be used to solve intractable games~\cite{soumalias2025llm,liu2025interpretable} and generalize classic game models to more realistic settings~\cite{fish2024generative,lu2024eliciting}, offering new computational approaches to complex game-theoretic problems.
\end{itemize}

\input{sections_journal/figures/_taxonomy}

Existing surveys on the intersection of game theory and LLMs primarily examine how game theory can be used to build evaluation environments and assess LLMs' strategic performance~\cite{zhang2024llm, feng2024survey, hu2024survey}. 
For instance, Zhang et al.~\cite{zhang2024llm} classified studies based on the game scenarios used to test LLM capabilities and methods for improving their reasoning. 
Meanwhile, Feng et al.~\cite{feng2024survey} and Hu et al.~\cite{hu2024survey} categorized the core competencies required for LLM-based agents in games, such as perception, memory, role-playing, and reasoning. 
These surveys offer valuable insights into the burgeoning field of LLM evaluation within strategic contexts. 
However, they predominantly adopt a unidirectional perspective, treating game theory as a tool for evaluating LLMs and improving LLMs' reasoning in games, overlooking the broader roles that game theory plays in developing LLMs and how LLMs are influencing game theory.  
This paper bridges this gap by \emph{introducing a novel four-part taxonomy} presented visually in Figure~\ref{fig:taxonomy}, which offers a holistic understanding of the synergistic interplay between these two critical domains. 
To our knowledge, this taxonomy provides \emph{the first truly comprehensive and structured analysis of the bidirectional relationship for this dynamic and interdisciplinary landscape.}
Therefore, we believe this work offers a nuanced and expansive perspective, enriching both domains.

%% file: sections_journal/figures/_taxonomy.tex
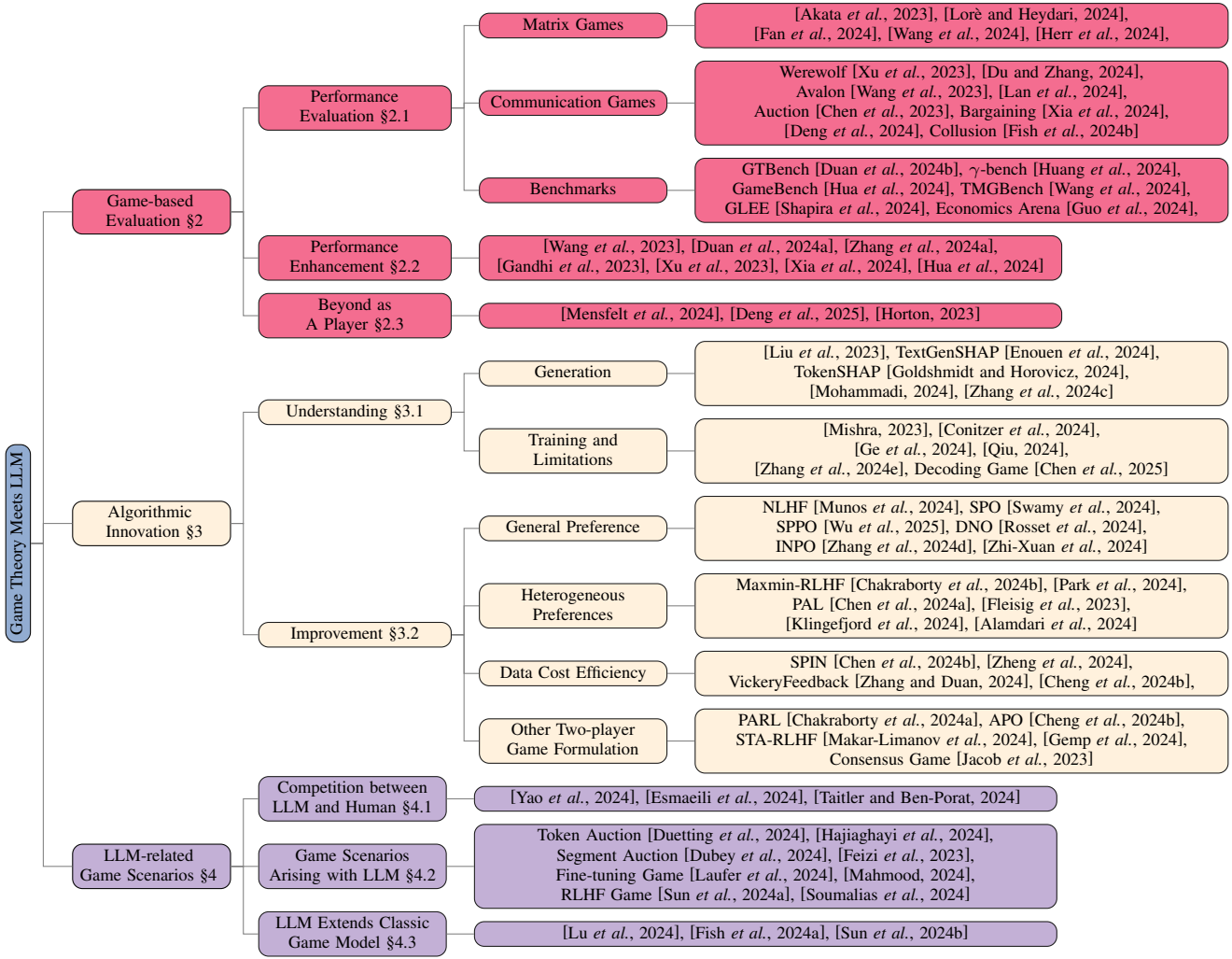
\begin{figure*}[t]
\centering
\resizebox{0.93\textwidth}{!}{%
\begin{forest}
  % --- 定义整个树的全局样式 ---
  for tree={
    grow'=east,                  % 从左向右生长
    forked edge,                 % "Y"形分叉线
    l sep=9mm,                  % 层级之间的距离
    s sep=5mm,                   % 同级节点之间的距离
    anchor=west,                 % 节点的锚点在左侧中间
    font=\sffamily,              % 使用无衬线字体
    rounded corners=3pt,         % 节点的圆角半径
    draw=edgecolor,              % 边框颜色
    edge={-latex, color=edgecolor, thick}, % 边的样式 (带箭头，加粗)
    line width=1pt,              % 边框线宽
    align=center,
  }
  % --- 根节点 (Survey 标题) ---
  [Game Theory Meets \\Large Language Models, fill=rootcolor, text=roottext, font=\sffamily\bfseries
    % --- Section 1 ---
    [Evaluating LLMs in \\Game-based Playground \S\ref{sec:game_for_llm_evaluation}, fill=level1color
      [Behavioral Features \S\ref{sec:game_for_llm_evaluation-observation}, fill=level2color]
      [Strategized Agents \S\ref{sec:game_for_llm_evaluation-strategizing}, fill=level2color]
      % [Benchmarks \S\ref{sec:game_for_llm_evaluation-benchmarks}, fill=level2color]
    ]
    % --- Section 2 ---
    [Improving LLMs with \\Game-theoretical Methods \S\ref{sec:game_for_llm_algorithm}, fill=level1color
      [Model Interpretability \S\ref{sec:game_for_llm_algorithm-interpretability}, fill=level2color]
      [General Preference \S\ref{sec:game_for_llm_algorithm-general}, fill=level2color]
      [Heterogeneity \S\ref{sec:game_for_llm_algorithm-heterogeneity}, fill=level2color]
      [Dynamic Adaptation \S\ref{sec:game_for_llm_algorithm-dynamic}, fill=level2color]
    ]
    % --- Section 3 ---
    [Characterizing LLM-related \\Events through Game Models \S\ref{sec:game_models}, fill=level1color
      [Multi-Stakeholder Competition \S\ref{sec:game_models-development}, fill=level2color]
      [Societal Impact of LLMs \S\ref{sec:game_models-implication}, fill=level2color]
    ]
    % --- Section 4 ---
    [Advancing Game \\Theory with LLMs \S\ref{sec:llm_for_game}, fill=level1color
      [Solving Intractable Games \S\ref{sec:llm_for_game-expand}, fill=level2color]
      [Expand Game-modeling \S\ref{sec:llm_for_game-solve}, fill=level2color]
    ]
  ]
\end{forest}
}
\caption{A taxonomy of the intersection between game theory and Large Language Models.}
\label{fig:taxonomy}
\Description{taxonomy}
\end{figure*}

%% file: sections_journal/2_game_for_llm_evaluation.tex
\section{Evaluating LLMs in Game-based Playground}\label{sec:game_for_llm_evaluation}
The integration of LLMs into game-based environments has emerged as a powerful avenue for evaluating their cognitive and decision-making capabilities. 
In contrast to traditional evaluation paradigms that emphasize linguistic competence, game-based evaluations reveal deeper dimensions of LLM behavior, such as human-like decision-making, uncertainty management, and strategic interaction. 
This section explores the growing role of games as playgrounds for LLM assessment: 
Subsection~\ref{sec:game_for_llm_evaluation-observation} analyzes the observable behaviors of LLMs when deployed as agents in games, highlighting decision-making patterns and interactive dynamics.
Subsection~\ref{sec:game_for_llm_evaluation-strategizing} focuses on the approaches to strengthen the strategic reasoning capabilities of LLMs, including prompting techniques, fine-tuning, and tool integration.
Together, these provide a comprehensive perspective on how games serve as both behavioral probes and enhancement platforms for LLMs.

\subsection{Observing Behavioral Features of LLM-based Agents}\label{sec:game_for_llm_evaluation-observation}

This subsection focuses on the behavioral characteristics exhibited by LLM-based agents in diverse game environments, capturing their decision-making tendencies, social interaction patterns, and responses to dynamic game states. 
By deploying LLMs in structured scenarios, ranging from matrix games and negotiation to deception and cooperation, researchers can uncover critical traits such as risk sensitivity, strategic adaptation, and behavioral consistency. 
These patterns not only reveal how LLMs interpret rules and model opponents but also expose cognitive limitations and human-like biases. A taxonomy of studies categorized by game types is presented in Table~\ref{tab:game_type_classification} to provide a structured overview of the empirical landscape.

\input{sections_journal/tables/2_game_categories}

\subsubsection{Basic Matrix Games}\label{sec:game_for_llm_evaluation-observation-matrix}
Matrix games represent foundational strategic settings in which players have complete information about the game structure, including all possible actions and payoffs. 
These games are often used as elementary tests of rational behavior~\cite{von2007theory}.
Recent studies employ natural language prompts to position LLMs as players in matrix games, instructing them to make decisions under various utility assumptions. 
Findings reveal that \emph{LLMs frequently exhibit pro-social biases, often prioritizing fairness and cooperation over game-theoretic rationality}. For instance, LLMs consistently display higher cooperation rates than humans in social dilemmas such as the Dictator Game~\cite{guo2023gptgametheoryexperiments, brookins2024playing, fontana2025nice, willis2025will}, and often reject unfair offers in the Ultimatum Game due to inequity aversion~\cite{horton2023large, ross2024llm, zheng2025nashequilibriumboundedrationality}.
The Turing Test conducted by Mei et al.~\cite{mei2024turing} also confirms that LLMs tend to behave more altruistically and cooperatively, which suggests that LLMs are maximizing the average of their and the partner's payoff by default. 
These behaviors likely stem from human-aligned moral patterns encoded during pre-training~\cite{vidler-walsh2025playing, jia2025largelanguagemodelstrategic}. 

At the same time, \emph{LLMs often deviate from optimality}, especially in tasks demanding probabilistic reasoning or adaptive play. 
For example, in zero-sum games like Rock-Paper-Scissors, many models fail to approximate mixed-strategy Nash equilibria~\cite{vidler-walsh2025playing, wu2024smartplay, silva2024largelanguagemodelsplaying}.
Repeated games like the Battle of the Sexes further reveal prompt sensitivity and coordination fragility~\cite{akata2023playing, duan2024reta, fan2024can, lore2024strategic}. 
In adversarial contexts, LLMs often revert to risk-averse or heuristic strategies~\cite{chen2024llmarena, gandhi2023strategic} and display bounded rationality that favors symmetric or fair outcomes over payoff maximization~\cite{zheng2025nashequilibriumboundedrationality, ross2024llm}.
Beyond specific games, systematic benchmarks such as FAIRGAME~\cite{buscemi2025fairgameframeworkaiagents} and SmartPlay~\cite{wu2024smartplay} also highlight the sensitivity of LLM behavior to prompt framing and contextual cues.

\subsubsection{Identity Games}\label{sec:game_for_llm_evaluation-observation-identity}
Identity games like Avalon, Werewolf, and Jubensha involve hidden roles, incomplete information, and social deception, making them rich environments for evaluating higher-order reasoning and strategic communication. 
In these games, agents must navigate uncertainty, infer hidden roles, and engage in deception, persuasion, and collaboration. Research in this area highlights a key duality, like LLM behavior.
Several studies employ LLM-based agents in an identity game-playground, letting different agents compete in the same game. 
Through thorough experimental observations, researchers find that \emph{LLMs can engage in recursive reasoning and social modeling.} 
For example, in Avalon, agents simulate others' beliefs and behaviors effectively~\cite{wang2023avalon}, while in Werewolf, some LLMs serve as ``opinion leaders,'' influencing teammates via persuasive summarization~\cite{du2024helmsman}.
Narrative games like Jubensha highlight LLMs' abilities to process long-form clues, infer roles, and construct coherent narratives~\cite{wu2024deciphering}.
On the other hand, \emph{LLMs often exhibit strategic inconsistencies and fragility}, particularly when faced with adversarial conditions or the need for robust logical deduction. AvalonBench reveals that current models have notable gaps in maintaining a consistent strategy and fully adapting to their assigned roles, especially when pressured by opponents~\cite{avalonbench}. In deduction-focused games like Werewolf, LLMs are prone to hallucination and flawed logical inference without external guidance~\cite{watanabe2024werewolf, bailis2024werewolfarena}. 
Despite appearing socially adept, these models remain susceptible to hallucinations, premature inferences, and are less aware of other players' intentions in high-stakes interactions~\cite{liu2024interintent}.

\subsubsection{Negotiation and Coordination Games}\label{sec:game_for_llm_evaluation-observation-negotiation}

Communication-based games, such as bargaining, provide a lens into LLMs' negotiation acumen and collaborative reasoning.
Studies show agents employing recognizable strategies like bluffing, anchoring, and making concessions during negotiations~\cite{zhan2024let, fu2023improvinglanguagemodelnegotiation, agashe2025llm}. 
More advanced models like GPT-4 are proficient at goal-directed planning and successfully making deals~\cite{xia2024measuring, abdelnabi2024llmdeliberation, qiao2023gameeval}. 
In contrast, less capable models often fail to maintain a consistent persona or even complete the negotiation task~\cite{yang2024reasoning, mukobi2023welfare, guan2025richelieu}. 
Meanwhile, their behavior remains highly sensitive to prompt phrasing, revealing a mixture of rational optimization and socially biased responses characteristic of bounded rationality~\cite{orner2025explaining, sreedhar2024simulatinghumanstrategicbehavior}.
In teamwork tasks, \emph{LLMs demonstrate emerging Theory of Mind (ToM) but struggle to maintain coordination in complex, belief-intensive environments.} 
While LLMs show a basic ability to infer the goals and intentions of teammates in games like Overcooked and Hanabi, their capacity for robust joint planning is limited~\cite{yim2024evaluatingenhancingllmsagent, agashe2025llm, piatti2024cooperate}. 
Davidson et al.~\cite{davidson2024evaluating} showed that cooperative bargaining games are the most challenging for LLMs. 
More specifically, models often regress to selfish or inconsistent strategies when social scaffolding is absent~\cite{xu2024magic, phelps2024machinepsychologycooperationgpt} and there are persistent challenges related to memory, deception, and adaptation, particularly in non-cooperative or socially complex contexts~\cite{qiao2023gameeval, xu2024magic, yang2024reasoning}. 

\subsubsection{Economic Games}\label{sec:game_for_llm_evaluation-observation-economics}
Economic games simulate market dynamics, requiring agents to make optimal decisions about pricing, bidding, and resource allocation under competitive pressure. 
Success in these environments demands a nuanced understanding of market forces, risk assessment, and the ability to anticipate and respond to competitors' actions. 
In these scenarios, including pricing games and auctions, \emph{LLMs demonstrate adaptive and sophisticated economic strategies, yet their performance is consistently bounded by imperfect reasoning and risk assessment.}
On the one hand, LLMs can emulate complex economic behaviors. 
For instance, tacit collusion and reward-punishment strategies emerge in repeated Bertrand games, with GPT-4 agents learning to maintain supracompetitive prices~\cite{fish2024algorithmic}. 
In dynamic auction environments, LLMs also show strategic planning by updating beliefs and reprioritizing items based on evolving conditions~\cite{chen2023put, guo2024economics}. 
However, this strategic acuity is often fragile and lacks robustness. Studies show that simpler heuristics can outperform LLMs, and equilibrium strategies are inconsistently achieved, revealing gaps in their strategic depth~\cite{chen2023put, guo2024economics, chen2024llmarena}. These observations reinforce the framing of LLMs as boundedly rational economic agents whose strategic capabilities are often superficial~\cite{immorlica2024generative}.

\subsubsection{Board and Card Games}\label{sec:game_for_llm_evaluation-observation-board}
Classic board and card games serve as demanding benchmarks for AI, as they require capabilities that are distinct from natural language processing. 
Success in these domains hinges on deep strategic planning, precise calculation, and sophisticated management of uncertainty. 
Across both perfect and imperfect-information games, \emph{baseline LLMs exhibit significant strategic deficiencies, struggling with deep calculation, logical consistency, and the management of uncertainty.}
In perfect-information games like Chess and Go, which demand rigorous, forward-looking planning, LLMs often fail to produce strategically coherent or even legally valid sequences of moves—a core deficiency that has necessitated the creation of specialized, search-augmented models to achieve competence~\cite{feng2023chessgpt, wang2025empowering}. This challenge extends to imperfect-information games, where the core difficulty shifts from pure calculation to managing strategic uncertainty. In poker, for instance, benchmarks show that vanilla LLMs perform poorly without significant fine-tuning, primarily due to their inability to handle hidden information and assess risk effectively~\cite{zhuang2025pokerbench, huang2024poker}. Similarly, in complex tactical battle games, these models are prone to hallucination and erratic play~\cite{hu2024pokellmon}. Taken together, evidence from multi-game testbeds confirms that deep planning limitations and strategic inconsistencies are pervasive weaknesses of general-purpose LLMs in these calculation-intensive contexts~\cite{chen2024llmarena}.

\subsubsection{Established Benchmarks}

To systematically probe the strategic capabilities of LLMs, a growing number of specialized benchmarks have been developed. These platforms provide controlled environments to assess specific facets of agent behavior, from pure rationality to complex social interaction. Several benchmarks focus on foundational strategic reasoning in classical games. For example, GTBench~\cite{duan2024gtbench}, $\gamma$-bench~\cite{huang2024far}, GameBench~\cite{hua2024game}, and the work by Topsakal et al.~\cite{grid-based-game-compete} primarily assess whether LLM decisions align with theoretical equilibria. Building on this, other frameworks like FAIRGAME~\cite{buscemi2025fairgameframeworkaiagents} and SmartPlay~\cite{wu2024smartplay} investigate the nuances of these decisions, examining deviations such as fairness biases, prompt sensitivity, and other signatures of bounded rationality.

The evaluation landscape has also expanded to encompass more dynamic and socially complex scenarios. Benchmarks like LLMArena~\cite{lan2024llm}, AvalonBench~\cite{avalonbench}, and NegotiationArena~\cite{bianchi24well} place LLMs in multi-agent settings to test higher-order recursive reasoning, coordination, and deception. The scope further extends into economic and hybrid domains, with GLEE~\cite{shapira2024glee} and TMGBench~\cite{wang2024tmgbench} evaluating agents in pricing and auction tasks, while ALYMPICS~\cite{mao2024alympicsllmagentsmeet} and Welfare Diplomacy~\cite{mukobi2023welfare} integrate negotiation and moral reasoning. To address concerns about agent fragility, recent additions like lmgame-Bench~\cite{hu2025imgamebench} and Playing Games~\cite{vidler-walsh2025playing} introduce metrics for disqualification and randomness-handling to better identify brittle behaviors. Collectively, this diverse and rapidly evolving suite of benchmarks provides an essential toolkit for diagnosing, comparing, and ultimately improving the strategic capabilities of LLM-based agents.

\subsection{Strategizing LLMs' Reasoning in Games}\label{sec:game_for_llm_evaluation-strategizing}

This subsection explores methods to enhance LLMs' strategic reasoning and performance in game-based environments, addressing the challenges of optimizing their decision-making processes. Techniques such as fine-tuning on game-specific data, incorporating reinforcement learning, or integrating external reasoning frameworks enable LLMs to better navigate complex game scenarios, from perfect-information games like chess to imperfect-information settings like poker. These approaches aim to improve LLMs' ability to anticipate opponents' moves, optimize strategies, and adapt to dynamic game states. This subsection highlights the advancements in enabling LLMs to exhibit sophisticated, goal-oriented reasoning in competitive and cooperative game contexts.

\subsubsection{Stimulating Reasoning with Advanced Prompting}

Advanced prompt techniques are often used to improve LLMs' reasoning capability~\cite{kojima2022large, wang2023selfconsistency, yao2023tree}. 
Several studies have designed more complex prompts for specific game tasks. 
Wang et al.~\cite{wang2023avalon} developed the Recursive Contemplation (ReCon) framework to enhance the strategic reasoning of LLMs in Avalon. By prompting LLMs to employ first- and second-order perspective-taking, this framework mitigates common failures like deceptive behavior.
Similarly, Duan et al.~\cite{duan2024reta} proposed a method where LLMs predict future moves in multi-turn games, improving their ability to anticipate opponents' strategies.
Additionally, Zhang et al.~\cite{zhang2025k} advanced LLMs' reasoning through $K$-level rationality, which enhances multi-level thinking and significantly increases their win rates in competitive settings. These findings suggest that recursive reasoning can substantially improve LLMs' strategic capabilities.
Kempinski et al.~\cite{kempinski2025got} proposed algorithms that guide LLMs to iteratively refine their action choices by simulating game outcomes in self-play.
These methods align directly with the themes discussed in this section, such as recursive reasoning and advanced prompting techniques for strategic capabilities.
Beyond recursive approaches, advanced prompting techniques also focus on integrating feedback, human-like reasoning, and Theory of Mind. 
Fu et al.~\cite{fu2023improvinglanguagemodelnegotiation} demonstrated that LLMs can autonomously improve negotiation strategies through self-play, leveraging in-context learning from AI feedback where a critic LLM provides structured critiques to a player LLM. 
In the context of multi-agent mystery games, Wu et al.~\cite{wu2024deciphering} enhanced agents' information gathering and logical reasoning by incorporating advanced prompting engineering, allowing them to decipher complex scenarios more effectively. 
Building on this, Guo et al.~\cite{guo2024suspicionagentplayingimperfectinformation} introduced ``Suspicion-Agent,'' which utilizes prompt engineering to harness GPT-4's high-order Theory of Mind capabilities, enabling it to understand and intentionally influence opponents' behavior in imperfect information games. Furthermore, Abdelnabi et al.~\cite{abdelnabi2024llmdeliberation} employed systematic zero-shot Chain-of-Thought (CoT) prompting to enable LLM agents to successfully negotiate in multi-agent games, highlighting the role of explicit reasoning steps. In a similar vein, Yim et al.~\cite{yim2024evaluatingenhancingllmsagent} proposed a ToM planning technique for LLM agents to adapt their strategies in cooperative games under imperfect information, demonstrating how specific prompts can simulate an understanding of other agents' beliefs. Lastly, Gandhi et al.~\cite{gandhi2023strategic} showed that systematically generated few-shot CoT examples can enable LLMs to achieve strategic reasoning that generalizes across diverse game structures and objectives. These diverse methods underscore the power of carefully designed prompts in stimulating sophisticated reasoning in LLMs across various game settings.

\subsubsection{Developing Task-Specific Ability with Training}

Complementing cognitive frameworks, novel training paradigms leverage self-play and AI feedback to overcome data limitations and improve strategic adaptability. Fu et al.~\cite{fu2023improvinglanguagemodelnegotiation} employed iterative self-play with AI-generated feedback to refine negotiation strategies in dynamic environments with hidden goals. 
Guo et al.~\cite{guo2024largelanguagemodelsplay} introduced self-supervised learning with auxiliary state-derived rewards, enabling mastery of complex games like Hanabi without human data. 
Kwon et al.~\cite{kwon2023rewarddesignlanguagemodels} used LLMs as intrinsic reward designers, reducing dependency on human-engineered reward functions for reinforcement learning. Complementing these, Zhang et al.~\cite{zhang2024agent-pro} implemented policy-level reflection via evolutionary algorithms, enabling LLM agents to self-optimize strategies without parameter retraining. Similarly, Wang et al.~\cite{wang2025empowering} tackled data scarcity in sensitive domains through algorithmic synthesis of statistically faithful game-theoretic scenarios using non-parametric copula simulators. These methods collectively enhance LLMs' ability to develop robust strategies through experiential learning.

Further advancements include an LLM-based framework by Wei et al.~\cite{wei2025automated}, which automates reward function discovery for reinforcement learning in cooperative platoon coordination, initializing rewards via a chain of thought and iteratively optimizing them through an evolutionary module based on training feedback. 
Suzuki and Arita~\cite{suzuki2024evolutionary} proposed an evolutionary model where LLMs are instructed with high-level psychological and cognitive character descriptions as ``genes'' to simulate human behavior in game-theoretical scenarios, evolving the population through selection based on average payoff and mutation of these linguistic trait descriptions. 
Feng et al.~\cite{feng2023chessgpt} introduced ChessGPT, which bridges policy learning and language modeling by integrating historical policy data and natural language analytical insights from chess games, training models on this large-scale combined dataset. 
Liao et al.~\cite{liao2024efficacy} demonstrated the efficacy of language model self-play in non-zero-sum games by fine-tuning LLMs over multiple rounds of filtered behavior cloning, showing substantial improvements in task reward. 
Wang et al.~\cite{wang2025empowering} empowered LLMs in decision games through targeted post-training by designing data synthesis strategies to curate extensive offline datasets from games like Doudizhu and Go, then developing techniques to effectively incorporate this data into LLM training. 
Jin et al.~\cite{jin2024learning} proposed an RL-instructed language agent framework for One Night Ultimate Werewolf, where a discussion policy is trained by reinforcement learning to determine strategic discussion tactics, guiding the LLM's communication based on game context. Zhuang et al.~\cite{zhuang2025pokerbench} introduced POKERBENCH, a benchmark for evaluating poker-playing abilities, demonstrating marked improvements in LLM performance after fine-tuning using structured ``Few-Shot Prompts'' that provide detailed game scenarios for strategic decision-making. 
Huang et al.~\cite{huang2024poker} presented PokerGPT, an end-to-end solver for multi-player Texas Hold'em that fine-tunes a lightweight LLM using reinforcement learning from human feedback based on textual records from real games. Finally, Yang and Berthellemy~\cite{yang2024reasoning} enhanced LLMs in non-cooperative games by integrating a tree of thoughts and a multi-agent framework, where game-solving is decomposed into incremental tasks and an automated fine-tuning process optimizes performance by ranking query-response pairs based on game outcomes.

\subsubsection{Integrating Auxiliary Modules and Tools}

Beyond direct prompting and training, integrating auxiliary modules and external tools is crucial for enhancing LLMs' game-playing ability by providing structured knowledge or specialized reasoning~\cite{schick2023toolformer, qin2024toolllm}. 
For instance, Yim et al.~\cite{yim2024evaluatingenhancingllmsagent} integrated a Theory of Mind planning technique with an external tool in Guandan, using prompts for strategic adaptation based on game context. Similarly, Xia et al.~\cite{xia2024measuring} enhanced bargaining with OG-Narrator, which employs prompts to structure offers and translate them into natural language.
Several works focus on infusing logical or strategic frameworks: Watanabe et al.~\cite{watanabe2024werewolf} improved Werewolf agents by embedding explicit logical structures via prompts for deductive reasoning. Wu et al.~\cite{wu2024deciphering} used advanced prompting within a multi-agent framework to boost information gathering and logical reasoning in mystery games. Hua et al.~\cite{hua2024game} developed a game-theoretic agent workflow for negotiation, guiding LLM decisions with specific prompts based on game theory. Lan et al.~\cite{lan2024llm} utilized a multi-agent system for Avalon, where the system prompts directed agents' gameplay and social behavior analysis.
Other approaches include Hu et al.~\cite{hu2024pokellmon}'s POKÉLLMON, which uses prompts to enable in-context reinforcement learning and knowledge-augmented generation for Pokémon battles. Guan et al.~\cite{guan2025richelieu} enhanced AI Diplomacy agents with a strategic planner that specifies sub-goals for long-term objectives. Lastly, the STRIDE framework~\cite{li2024stride} integrates memory and specialized tools, with prompts enabling LLM agents to interact for rule adherence, planning, exploration, and opponent anticipation. These diverse integrations significantly bolster LLMs' strategic capabilities.

\begin{tcolorbox}[
    colback=white!5!white,      % 方框背景颜色 (这里是浅灰色)
    colframe=gray!75!black,     % 方框边框颜色 (这里是深灰色)
    title=\textbf{Discussions}, % 方框标题，加粗
    fonttitle=\bfseries,        % 标题字体加粗
    boxsep=5pt,                 % 方框内容与边框的距离
    arc=5pt,                    % 圆角半径
    outer arc=5pt,              % 外圆角半径
    left=5pt,right=5pt,top=5pt,bottom=5pt % 内边距
]
Many current findings on the behavioral characteristics of LLMs are closely tied to specific model architectures or versions, making them potentially obsolete as the technology rapidly evolves. It would be highly valuable to explore or derive more fundamental and generalizable strategic patterns of LLM behavior through game-theoretic scenarios that transcend individual models. While recent efforts to enhance LLMs' reasoning abilities have largely focused on task-specific approaches, developing a unified framework for improving general game-playing and reasoning capabilities remains a significant and open challenge.
\end{tcolorbox}

The observed behavioral patterns, particularly the pro-social biases and strategic limitations, are not arbitrary phenomena. They are, in large part, artifacts of the very training and alignment techniques used to build these models. The tendency towards fairness and cooperation, for example, is a direct result of methods designed to make LLMs helpful and harmless. In the next section, we turn our attention to these underlying mechanisms, examining how game-theoretic principles are themselves being used to improve LLMs, which in turn shape the behaviors we have just reviewed.

%% file: sections_journal/tables/2_game_categories.tex
\begin{table}[t]
\caption{A Summary of LLM-based Agents' Behavioral Features in Various Game Categories.}
\label{tab:game_type_classification}
\centering
\resizebox{0.88\textwidth}{!}{%
\begin{tabularx}{\textwidth}{@{} >{\raggedright\arraybackslash}X >{\raggedright\arraybackslash}p{2.7cm} >{\raggedright\arraybackslash}p{2.1cm} >{\raggedright\arraybackslash}p{2.1cm} >{\raggedright\arraybackslash}p{2.1cm} @{}}
\toprule
\textbf{Game Category} & \textbf{\makecell{Information}} & \textbf{\makecell{Nature}} & \textbf{\makecell{Interaction}} & \textbf{\makecell{Structure}} \\
\midrule

\textbf{Basic Matrix Games}~\S\ref{sec:game_for_llm_evaluation-observation-matrix} \newline \emph{e.g.}, Prisoner's Dilemma, Ultimatum Game, RPS \cite{guo2023gptgametheoryexperiments, horton2023large, zheng2025nashequilibriumboundedrationality, vidler-walsh2025playing, akata2023playing} & Perfect / Imperfect & Comp. / Coop. & Single-turn / Repeated & Symmetric / Asymmetric \\
\addlinespace

\rowcolor{lightgray}
\multicolumn{5}{@{}p{\dimexpr\textwidth-2\tabcolsep}}{
\textit{Spotlight:} LLMs exhibit strong pro-social biases (\emph{e.g.,} fairness, cooperation) often deviating from game-theoretic rationality. They struggle with probabilistic reasoning and approximating mixed-strategy Nash equilibria, showing high sensitivity to prompt framing.} \\
\midrule

\textbf{Identity Games}~\S\ref{sec:game_for_llm_evaluation-observation-identity} \newline {\emph{e.g.}, Avalon, Werewolf, Jubensha \cite{wang2023avalon, du2024helmsman, wu2024deciphering, avalonbench}} & Imperfect & Comp. / Coop. & Multi-turn, Social & Asymmetric \\
\addlinespace

\rowcolor{lightgray}
\multicolumn{5}{@{}p{\dimexpr\textwidth-2\tabcolsep}}{
\textit{Spotlight:} Capable of recursive reasoning and social modeling (\emph{e.g.,} influencing teammates). However, they lack strategic reliability, failing to maintain role consistency and logical coherence under pressure, and are prone to hallucinations.} \\
\midrule

\textbf{Negotiation \& Coordination}~\S\ref{sec:game_for_llm_evaluation-observation-negotiation} \newline {\emph{e.g.}, Bargaining, Overcooked, Hanabi \cite{zhan2024let, xia2024measuring, yim2024evaluatingenhancingllmsagent, piatti2024cooperate, davidson2024evaluating}} & Imperfect & Comp. / Coop. & Multi-turn, Communicative & Symmetric / Asymmetric \\
\addlinespace

\rowcolor{lightgray}
\multicolumn{5}{@{}p{\dimexpr\textwidth-2\tabcolsep}}{
\textit{Spotlight:} Demonstrate recognizable negotiation tactics (bluffing, anchoring) and emerging Theory of Mind (ToM). Yet, robust coordination is limited; they often regress to selfish or inconsistent strategies in complex scenarios without social scaffolding.} \\
\midrule

\textbf{Economic Games}~\S\ref{sec:game_for_llm_evaluation-observation-economics} \newline {\emph{e.g.}, Bertrand Competition, Auctions \cite{fish2024algorithmic, chen2023put, guo2024economics, immorlica2024generative}} & Imperfect & Comp. & Multi-turn / Repeated & Symmetric \\
\addlinespace

\rowcolor{lightgray}
\multicolumn{5}{@{}p{\dimexpr\textwidth-2\tabcolsep}}{
\textit{Spotlight:} Display adaptive economic strategies, such as tacit collusion in pricing games. Performance is bounded by imperfect reasoning and fragile risk assessment, often failing to achieve equilibrium strategies consistently.} \\
\midrule

\textbf{Board \& Card Games}~\S\ref{sec:game_for_llm_evaluation-observation-board} \newline {\emph{e.g.}, Chess, Go, Poker \cite{feng2023chessgpt, zhuang2025pokerbench, hu2024pokellmon, chen2024llmarena}} & Perfect / Imperfect & Comp. & Multi-turn & Symmetric \\
\addlinespace

\rowcolor{lightgray}
\multicolumn{5}{@{}p{\dimexpr\textwidth-2\tabcolsep}}{
\textit{Spotlight:} Baseline models show significant strategic deficiencies. They struggle with deep calculation, logical consistency, and managing uncertainty information, often producing invalid or strategically incoherent moves without specialized fine-tuning.} \\

\bottomrule
\end{tabularx}%
}
\end{table}

%% file: sections_journal/3_game_for_llm_algorithms.tex
\section{Improving LLMs with Game-theoretical Methods}\label{sec:game_for_llm_algorithm}

Game-theoretical approaches have been more frequently utilized to describe LLMs' theoretical characteristics and to develop practical algorithms that improve their empirical outcomes. 
This section explores how principles and methodologies from game theory contribute to addressing key challenges in the development and optimization of LLMs.
The discussion is then organized into four subsections, each corresponding to a critical challenge faced by LLMs:
Subsection~\ref{sec:game_for_llm_algorithm-interpretability} addresses the challenge of \emph{interpretability}. A line of work constructs cooperative game scenarios involving an LLM's input, training data, and internal components, and utilizes the \emph{Shapley value}~\cite{shapley1953value} to provide principled credit assignment for each contributing factor.
Subsection~\ref{sec:game_for_llm_algorithm-general} focuses on \emph{aligning general preferences}. Nash Learning from Human Feedback (NLHF) formulates a \emph{minimax game between policies} to capture intransitive preferences. Further extensions develop more efficient algorithms with provable theoretical guarantees and broader applicability. 
Subsection~\ref{sec:game_for_llm_algorithm-heterogeneity} discusses the challenge of \emph{heterogeneity} in value alignment. 
Recent work integrates \emph{social choice theory} with reinforcement learning from human feedback (RLHF), offering axiomatic frameworks for alignment and principled strategies to resolve diverse, potentially conflicting preferences.
Subsection~\ref{sec:game_for_llm_algorithm-dynamic} investigates the issue of \emph{dynamic adaptation} by modeling LLM development as a \emph{competitive game} involving multiple co-evolving players. 
Table~\ref{tab:game_for_llm_improvement} summarizes the key challenges, core game-theoretical concepts, and corresponding game formulations.

\input{sections_journal/tables/3_game_for_llm_improvement}

\subsection{Enhancing Model Interpretability through Cooperative Game}\label{sec:game_for_llm_algorithm-interpretability}

Interpreting the behavior of deep learning models is a highly desirable objective, as it enables researchers to gain insights into how models function, thereby facilitating further improvements and providing guarantees of reliability~\cite{chakraborty2017interpretability, zhang2018visual, poursabzi2021manipulating}. Such an interpretability challenge becomes more critical for LLMs, due to their massive scale, often comprising billions or even trillions of parameters and trained on vast datasets.
The SHAP framework~\cite{lundberg2017unified} adopts the Shapley value~\cite{shapley1953value} to quantify the contribution of individual components of a deep learning model. 
Inspired by this, recent work considers formulating cooperative games in different stages of LLMs, in which each player represents a semantically meaningful part, such as individual tokens in a prompt, specific datasets used in training, or particular layers within the LLM architecture. 
By choosing appropriate performance metrics, the researchers apply the Shapley value to measure how each player contributes to the overall performance, offering a theoretically grounded measure of interpretability. 
However, the exponential complexity of exact Shapley value calculation necessitates the development of efficient and accurate approximation techniques, which has become a central theme in this line of research.

\subsubsection{Input Attribution}

Several studies focus on quantifying the importance of an LLM's input components, from entire prompts to individual tokens. 
At the prompt level, Liu et al.~\cite{liu2023prompt} applied the Shapley value to multi-prompt learning. They proposed a learning-based approach to predict the value of each prompt, facilitating more effective prompt engineering by identifying prompts that enhance performance.
At a finer granularity, TokenSHAP~\cite{horovicz-goldshmidt2024tokenshap} and TextGenSHAP~\cite{enouen2023textgenshap} model individual tokens as players in a cooperative game to attribute the model's output to specific parts of the input. TextGenSHAP~\cite{enouen2023textgenshap} introduces speculative decoding and in-place encoder resampling to make token-level attribution for long-context inputs computationally tractable. 
Using a similar method, Mohammadi~\cite{mohammadi2024explaining} uncovered a ``token noise'' phenomenon: LLM decisions are disproportionately affected by tokens with minimal semantic content, like invisible newline characters. 
Moving from token-level to document-level analysis, Ye and Yoganarasimhan~\cite{ye2025document} applied the Shapley framework to value source documents in LLM-generated summaries, a key challenge in Retrieval-Augmented Generation systems. They introduced Cluster Shapley, which groups semantically similar documents, to overcome the computational expense.
In their model, each source document is a player whose value is determined by its marginal contribution to the quality of the final summary.

\subsubsection{Training Data Valuation}

As LLMs are trained on various massive datasets, it is insightful to quantify how each dataset or data instance influences the model's performance.
\emph{Data Shapley}~\cite{ghorbani2019data} models each training sample as a player and uses the Shapley value to evaluate its marginal contribution to the final model's performance across all possible training subsets. 
In the LLM context, this concept has been operationalized for both dataset refinement and dynamic learning. 
For dataset curation, the SHED framework~\cite{he2024shed} offers a scalable solution for instruction fine-tuning. It approximates Shapley values over clusters of training data to assemble an optimized dataset. 
The framework also demonstrates strong cross-model transferability, where data selected using a smaller model remains effective for larger ones, significantly reducing curation costs. 
Beyond static dataset curation, this principle extends to the dynamic process of RLHF. The SCAR framework~\cite{cao2025scar} addresses the sparse reward problem in RLHF by modeling text segments as players in a cooperative game. 
By distributing the final reward among text units based on their Shapley value—approximated efficiently using Owen values—SCAR provides a dense, principled reward signal. 
Empirically, this method improves both convergence speed and final policy performance in tasks like summarization and instruction following.
% Crucially, unlike attention-based methods, it can assign negative credit to detrimental text segments and is theoretically guaranteed to preserve the optimal policy, improving both convergence speed and final policy performance in tasks like summarization and instruction following.

\subsubsection{Probing Internal Components}

Beyond external inputs and data, Shapley-based methods are also applied to the internal components of LLMs. By conceptualizing architectural components such as layers and attention heads as ``players'' in a cooperative game, researchers can assess the marginal contribution of each part to the model's overall performance. This abstraction provides not only interpretability insights but also practical guidance for model optimization and compression.
One line of work evaluates the importance of entire layers. Zhang et al.~\cite{zhang2024investigating} used a Shapley-based neighborhood sampling method to identify \emph{cornerstone layers}, a small subset of layers whose removal causes a drastic performance collapse.
Following this, Sun et al.~\cite{sun2025efficient} proposed the Shapley Value-based Non-Uniform Pruning (SV-NUP) framework. It employs an efficient Sliding Window-based Shapley Value (SWSV) approximation to assign pruning ratios based on layer contributions, achieving superior compression-performance trade-offs. 
Other studies focus on finer-grained components like attention heads. 
Held and Yang~\cite{held-yang2023shapley} applied \emph{Shapley Head Values} (SHVs) to identify and remove interfering attention heads in multilingual models, achieving performance improvement without adding or retraining parameters. 
To make such fine-grained analysis computationally feasible, amortized methods have been introduced. 
Yang et al.~\cite{yang2023efficient} trained an auxiliary model to directly predict Shapley values, achieving speedups while ensuring deterministic, stable explanations.
Fekete and Bjerva~\cite{fekete2025linguistically} used SHVs and clustered the resulting value vectors to reveal how different attention heads specialize in processing specific morphosyntactic phenomena.
% Furthermore, frameworks like TransSHAP~\cite{kokalj2021bert} adapt SHAP for transformers, enabling sequential visualization of internal contributions. 
%  Collectively, these works establish Shapley-based analysis as a powerful and versatile tool for demystifying LLM internals.
Furthermore, Yang et al.~\cite{yang2025whosmvp} proposed a benchmark and evaluation metrics based on Shapley value to evaluate how each module, such as planning, reasoning, action execution, and reflection, contributes to the reasoning improvement of an LLM.  

\subsection{Aligning General Preferences through Min-Max Equilibrium}\label{sec:game_for_llm_algorithm-general}

Aligning LLMs with general human preferences is a core challenge in the deployment of responsible, reliable language models. 
Traditional RLHF typically relies on reward models built upon the Bradley-Terry (BT) assumption, which imposes a strict transitive and scalar reward structure~\cite{ouyang2022training, kaufmann2024surveyrlhf}. 
However, human preferences are often stochastic and intransitive, which are poorly modeled by BT-style assumptions~\cite{NLHF, SPO}. 
Recent research reconceptualizes alignment as a two-player min-max game between policies and proposes a series of novel algorithms based on Nash equilibrium, self-play, and preference optimization without explicit reward modeling. 
These approaches aim to directly capture complex, expressive feedback and offer both theoretical guarantees and empirical improvements.

\subsubsection{Preference Optimization}
The foundational idea of NLHF~\cite{NLHF} is to model alignment as a two-player zero-sum game, where each policy aims to outperform the other based on a learned preference function. This approach does not need reward modeling and allows convergence toward a Nash equilibrium policy under general preferences. 
Building on NLHF, Self-Play Preference Optimization (SPO)~\cite{SPO} proposes a minimalist yet generalizable framework. SPO treats trajectory comparison as a symmetric game and uses self-play to train a policy against itself using win-rate scores, achieving robustness to non-Markovian and intransitive preferences. 
Self-Play Preference Optimization for LLMs (SPPO)~\cite{SelfPlayPO} adapts SPO for practical LLM fine-tuning by introducing a new squared-loss objective. 
% This objective, motivated by policy gradient theory, directly approximates the ideal multiplicative weight update rule, implicitly encouraging the model to learn a token-level optimal value function without explicit reward modeling.
To address sample efficiency and training stability, Direct Nash Optimization (DNO)~\cite{DNO} introduces a batched, on-policy, regression-based objective. DNO combines the scalability of contrastive learning with Nash theoretical soundness and is more empirically efficient. 
% A key practical insight from DNO is the importance of leveraging preference pairs where the student model outperforms the teacher model, a mechanism crucial for enabling iterative self-improvement beyond the teacher's capabilities.
Similarly, Iterative Nash Policy Optimization (INPO)~\cite{IiterativeNPO} proposes a no-regret online learning framework that avoids costly win-rate estimation by directly minimizing a surrogate loss over a preference dataset. 
% INPO approximates the Nash policy while achieving strong empirical performance and stability in large-scale language model alignment tasks.
Furthering this direction, Ye et al.~\cite{ye2024online} provided a theoretical study of KL-regularized RLHF under a general preference oracle, completely bypassing the BT model. They formulate alignment as a minimax game and propose distinct algorithms for offline (Pessimistic Equilibrium Learning, PELHF) and online (Optimistic Equilibrium Learning, OELHF) settings, both with finite-sample theoretical guarantees. 
The learnability of the KL-regularized NLHF is further verified in the work by Ye et al.~\cite{ye2024theoretical}.
% Their work demonstrates that general preference models can achieve better generalization, especially on out-of-distribution reasoning tasks. 
% Critically, their theoretical analysis underscores the limitations of purely offline learning, which can require exponential data coverage, thus emphasizing the need for online data collection to ensure effective and sample-efficient policy optimization.

\subsubsection{Theoretical Advancements in Convergence and Efficiency}
Further studies are focusing on achieving provably efficient and convergent training dynamics for NLHF. 
Several recent works observe oscillatory behavior, high variance, and slow convergence in preference-based self-play.
Nash Mirror Prox (NashMP)~\cite{tiapkin2025accelerating} addresses these issues by leveraging the Mirror Prox method in online Nash Learning from Human Feedback. 
NashMP achieves last-iterate linear convergence to the $\beta$-regularized Nash equilibrium, with convergence rates that are independent of action space size. 
Extragradient Preference Optimization (EGPO)~\cite{zhou2025extragradient} extends this idea, providing both linear convergence for regularized games and polynomial convergence for unregularized settings.
% It avoids nested optimization loops common in other methods by deriving an equivalent online update scheme, improving both scalability and robustness to noisy gradients.
Magnetic Preference Optimization (MPO)~\cite{wang2024magnetic} is designed to converge to the Nash equilibrium of the \textit{original, non-regularized} game. It achieves this by periodically updating its reference ``magnetic'' policy to the equilibrium of the previous regularized step, guiding the policy sequence toward the unregularized objective.
Beyond theoretical convergence, Wang et al.~\cite{wang2024provably} proposed TANPO (Two-Agent Nash Policy Optimization) to improve sample efficiency and exploration-exploitation balancing. 
In TANPO, the ``min-player'' is incentivized to explore via an exploration bonus, thereby generating more diverse and informative training data for the ``max-player''. 
Other work stabilizes training through regularization. 
Tang et al.~\cite{tang2025game} introduced Regularized Self-Play Policy Optimization (RSPO) that integrates forward and reverse KL divergence regularization into self-play, allowing for tunable trade-offs between response length, win-rate, and diversity. 
% Empirically, they uncover distinct effects of different regularizers: forward KL divergence tends to shorten response length, while reverse KL divergence significantly boosts the raw win-rate, providing fine-grained control over model behavior. 
% This approach demonstrates significant performance gains over unregularized SPPO on standard alignment benchmarks, effectively mitigating the over-optimization issues observed in later training iterations.
Alami et al.~\cite{alami2024investigating} further explored the dynamics of KL-based regularization in self-play, showing that geometric mixing of the base and reference policies improves performance stability. 
They also highlight how fictitious play helps smooth the training trajectory and prevent performance oscillations, which helps prevent training collapse and boost generalization across benchmarks.

\subsubsection{Limitations of Game-Theoretic Alignment}
Although game-theoretic methods for alignment provide flexibility and robustness to preference inconsistencies, there are foundational critiques that raise concerns about the sufficiency of preferences for full value alignment.
Sun et al.~\cite{sun2024rethinking} revisited the foundational Bradley-Terry model, arguing that while it offers order consistency, it may not be necessary or optimal. They advocate for classification-based alternatives and emphasize the role of annotation sparsity and structure in reward modeling quality. Going deeper, Shi et al.~\cite{shi2025fundamental} investigated theoretical limits of game-theoretic alignment, demonstrating fundamental limits to preference matching. They proved that no smooth payoff mapping can guarantee an equilibrium solution that precisely matches a target preference profile, nor that such an equilibrium is unique. They further showed that achieving other desirable properties, such as Smith consistency, imposes strict structural constraints on the game, requiring it to be equivalent to a symmetric zero-sum game.
From a philosophical perspective, Zhi et al.~\cite{zhi2024beyond} challenged the entire preferentist paradigm. They argued that preferences lack the semantic richness and context-dependence of human values, and thus fail as alignment targets. They called for a reframing of alignment to be based on normative standards tailored to AI roles, negotiated across stakeholders, rather than preference aggregation.

\subsection{Capturing Preference Heterogeneity with Social Choice Theory}\label{sec:game_for_llm_algorithm-heterogeneity}

Another formidable challenge in aligning large language models with human preferences is the heterogeneity of human preferences themselves. Conventional alignment methods, which often rely on learning a single, scalar reward function, overly prefer the majority's point and fail to handle this heterogeneity~\cite{fleisig2023majority}. 
To address this problem, game theory and social choice theory offer both a formal language to diagnose the core difficulties of this problem and a constructive paradigm for developing more robust and equitable alignment algorithms.

\subsubsection{An Axiomatic Social Choice Perspective}

Recent work has recognized the limitations of RLHF by formally \emph{mapping the alignment problem to the field of social choice theory}. 
This perspective reveals that many challenges are not merely technical but are rooted in fundamental paradoxes of collective decision-making. Mishra~\cite{mishra2023ai} showed that preference aggregation in AI is subject to classic impossibility theorems from social choice. 
By mapping AI alignment to this setting, he demonstrated that any aggregation rule inevitably violates at least one of Arrow's axioms—such as unanimity, independence of irrelevant alternatives, or non-dictatorship—meaning no single ``fair'' preference aggregator can exist. 
Complementing this, Dai and Fleisig~\cite{dai2024mapping} formalized the mapping between RLHF and social choice, enabling a deeper critique. 
For example, it has been shown that canonical RLHF methods based on the Bradley-Terry model are mathematically analogous to the Borda count voting rule~\cite{shirali2025direct}, where an option's score is derived from the sum of its pairwise ``wins.'' 
This connection highlights how RLHF implicitly adopts a specific, and not always desirable, social choice function.
\emph{This axiomatic lens also provides a powerful tool for critiquing current alignment mechanisms by testing them against formal properties.}
Ge et al.~\cite{ge2024axioms} evaluated RLHF algorithms and found they violate fundamental axioms like Pareto Optimality and Pairwise Majority Consistency. 
Extending this critique, empirical work by Hosseini and Khanna~\cite{hosseini2025distributive} confirms this misalignment, showing that LLMs often default to welfare-maximization in resource allocation tasks, violating human concepts of distributive fairness like equitability.
Procaccia et al.~\cite{procaccia2025clone} identified a critical vulnerability to ``approximate clones''—semantically similar responses that trick MLE-based methods by artificially inflating an option's representation in the data; they showed these methods fail the axiom of clone-independence and propose a weighted MLE solution to correct it. 
Extending the critique to utility, Gölz et al.~\cite{golz2025distortion} introduced the concept of ``distortion'' as the worst-case ratio of optimal social welfare to the welfare achieved by an RLHF policy, proving that for methods like RLHF, this distortion can grow exponentially with preference intensity, and can even be unbounded under certain data sampling conditions.
In contrast, Xiao et al.~\cite{xiao2025theoretical} offered a reconciliation, explaining why these theoretical failures don't always break RLHF in practice. They demonstrated that under a common and empirically plausible condition—where each response pair is evaluated by at most one annotator—the complex preference cycles needed to trigger many axiomatic violations rarely form, thus preserving properties like Condorcet Consistency. 
This highlights a central tension in the field: the clash between theoretical impossibility results and the empirical effectiveness of existing methods under specific, practical conditions.

Building on these insights, \emph{researchers have proposed new axioms and normative frameworks to guide the development of more principled alignment systems}. 
Position papers by Conitzer et al.~\cite{conitzer2024position} and Zhang et al.~\cite{zhang2024incentive} argue for embedding principles from social choice and mechanism design directly into the alignment process. The latter introduces the Incentive Compatibility Sociotechnical Alignment Problem (ICSAP), which uses mechanism design to create protocols where stakeholders are incentivized to reveal their true preferences. 
This concern with strategic incentives is formalized by Wu et al.~\cite{wu2025battling}, whose game-theoretic analysis of competing data providers shows that strategic exaggeration of preferences is an almost inevitable Nash Equilibrium, highlighting a fundamental source of data corruption.
To build more legitimate systems, others are designing new elicitation processes, such as the ``Moral Graph'' proposed by Klingefjord et al.~\cite{klingefjord2024human}, which uses a structured dialogue to construct a graph of human values capable of surfacing nuanced and expert opinions.
Another direction focuses on axioms of representation. 
Kim et al.~\cite{kim2025population} introduced a framework grounded in two new axioms: Population-Proportional Representation (PPR) and Population-Bounded Robustness (PBR), achieved by learning a latent mixture of preference groups. This call for proportional representation is echoed by Peters~\cite{peters2024proportional}, who argues for its broad applicability in AI to mitigate the ``tyranny of the majority'' in contexts ranging from RLHF to the aggregation of LLM outputs.
Complementing this, Qiu~\cite{qiu2024representative} connects alignment to statistical learning theory, proposing axioms that ensure a preference model learned from a sample of users will generalize fairly to the entire population.

\subsubsection{Developing Practical Algorithms}

The theoretical shortcomings of a single, monolithic reward function, particularly its failure to satisfy axioms like Pareto Optimality~\cite{ge2024axioms} and its tendency towards high distortion~\cite{golz2025distortion}, motivate a new class of practical algorithms designed to explicitly handle preference diversity. Empirical work confirms this necessity. 
For example, Fleisig et al.~\cite{fleisig2023majority} treated annotator disagreement as a crucial signal, developing a model to predict the ratings of a text's targeted group to identify instances where the majority opinion is ``wrong''.
Shirali et al.~\cite{shirali2025direct} provided theoretical proof that Direct Preference Optimization (DPO) fails to find the utilitarian optimum with heterogeneous users, instead implicitly optimizing for the Borda count and thus being sensitive to data sampling distribution. This work also uncovers a trade-off: achieving a consistent estimate of the optimal policy requires discarding data with annotator disagreement, thereby sacrificing sample efficiency. 
Directly addressing this, Cheng et al.~\cite{cheng2024self} proposed Vote-based Preference Optimization (VPO), which leverages the quantitative vote counts in preference data. By modeling preference strength, VPO can distinguish between clear consensus and controversial opinions, leading to more stable and effective alignment.
This issue of inconsistency is underscored by Liu et al.~\cite{liu2025statistical}, who show that Condorcet cycles (\emph{e.g.,} A > B, B > C, C > A) are a near-certainty in diverse preference data. Since a single scalar reward function cannot represent such cycles, this proves its structural inadequacy and motivates algorithms capable of producing mixed or pluralistic outputs.

One major line of work involves learning distinct reward models for different preference clusters and then aggregating their outputs based on principles from social choice. 
Chakraborty et al.~\cite{chakraborty2024maxmin} proposed MaxMin-RLHF, which implements the Rawlsian ``max-min'' social welfare function by learning a mixture of reward models for latent subgroups and maximizing the utility of the worst-off group. Similarly, Chen et al.~\cite{chen2024pal} used a mixture-of-experts approach to learn latent preference prototypes, while Park et al.~\cite{park2024rlhf} used spectral clustering to identify taste-based groups before training group-specific reward models. Providing a theoretical backing, Zhong et al.~\cite{zhong2024provable} used meta-learning to extract group-specific rewards and analyze the sample complexity of aggregating them using different social welfare functions.

Other approaches innovate on the aggregation mechanism itself, moving beyond the ``learn-then-aggregate'' paradigm. Alamdari et al.~\cite{alamdari2024policy} proposed aggregating at the policy level, where multiple policies trained on individual preferences have their actions combined at inference time using voting rules. 
The viability of such voting mechanisms is empirically explored by Yang et al.~\cite{yang2024llm}, who find that LLM collective decisions are sensitive to voting protocols and exhibit biases. While using ``personas'' can improve alignment with human choices, it reveals a difficult trade-off between alignment accuracy and preference diversity.
Halpern et al.~\cite{halpern2025pairwise} introduced Pairwise Calibrated Rewards, an ensemble method that learns a distribution of reward functions calibrated to match the proportion of human annotators holding a given preference, thus preserving pluralism.
Drawing from cooperative game theory, Mushkani et al.~\cite{mushkani2025negotiative} proposed Negotiative Alignment, a multi-agent framework where agents representing stakeholder groups use bargaining protocols to reach collective decisions.
The potential of such negotiative approaches is highlighted by behavioral experiments from Wang et al.~\cite{wang2024large}, who show that an ``imperfectly fair'' LLM agent—one that engages in human-like strategic communication—can overcome the typical ``machine penalty'' and foster cooperation in social dilemmas, suggesting the power of dynamically negotiated alignment.

The integration of social choice theory has transformed the understanding of AI alignment with heterogeneous preferences, shifting the focus from a single reward function to a rich tapestry of methods that embrace diversity. Future work will likely focus on bridging the gap between theoretically ideal but computationally expensive mechanisms, such as those involving negotiation or full pairwise calibration, and scalable algorithms that can be deployed in real-world systems, while also addressing the crucial challenge of incentive compatibility in data collection.

\subsection{Achieving Dynamic Adaptation through Competitive Game}\label{sec:game_for_llm_algorithm-dynamic}

Traditional LLM training pipelines often rely on two fixed components: a static, pre-collected dataset for supervised fine-tuning, and a static reward model trained on human preferences. 
This static nature imposes fundamental limitations: a model's capabilities are constrained by the fixed data it has seen, and an evolving policy can learn to exploit loopholes in the fixed reward model, leading to a brittle alignment known as ``reward hacking''~\cite{skalse2022defining}.
Game theory provides a robust alternative by recasting LLM training and deployment as a dynamic, strategic, and interactive process.

\subsubsection{Overcoming Static Datasets}

Static datasets represent a finite and fixed data distribution, and collecting human-annotated data is expensive.
This restricts a model's ability to generalize to novel scenarios. 
To address this, a prominent line of research uses self-play, where an LLM iteratively generates its training data. 
The SPIN framework~\cite{SPIN}, for example, has the LLM play against previous versions of itself. In each round, the model generates new responses and learns to distinguish these from a seed set of human-annotated data, effectively creating a curriculum of progressively higher-quality data without needing continuous human feedback. 
This concept of self-alignment is also explored by Azarafrooz et al.~\cite{azarafrooz2024language}, who propose an online, two-player game that can be viewed as a simplified form of DPO operating without any human preference data, using Nash-learning and adaptive feedback to enable autonomous improvement. 
Similarly, Chu et al.~\cite{chu2025stackelberg} tackled the data scarcity and noise problem simultaneously with their Stackelberg Game Preference Optimization (SGPO). This framework uses self-annotation to create worst-case preference data efficiently, enhancing the robustness of the model.

Adversarial games are also used to specifically target and patch a model's weaknesses. In this paradigm, one agent's goal is to generate inputs that the other agent finds difficult. Zheng et al.~\cite{zheng2024toward} let an adversarial agent generate prompts that expose the weaknesses of a defensive agent. They also introduce an innovative diversity constraint to prevent the adversary from collapsing to a narrow set of attacks.
The utility of such adversarial dynamics is profound; for instance, self-play in an ``Adversarial Taboo'' game is more effective for enhancing LLM reasoning than standard supervised fine-tuning~\cite{cheng2024self}.
Ye et al.~\cite{ye2025reward} proposed a creator-solver dynamic where a ``creator'' model strategically crafts new prompts by aiming to maximize the ``solver'' model's regret. 
Similarly, Liu et al.~\cite{liu2025chasing} used online self-play to co-evolve attacker and defender agents with mechanisms like a ``Hidden Chain-of-Thought'' to enhance strategic planning. 
% The underlying self-play algorithms are also being refined; for example, by fusing historical policies according to a Nash equilibrium to accelerate convergence in the Policy Space Response Oracles (PSRO) framework~\cite{lian2024fusion}.

Another strategy involves using game dynamics not just to generate prompts, but to create fine-grained, step-by-step feedback, which is notoriously difficult to obtain from humans. Chen et al.~\cite{chen2025spc} introduced the Self-Play Critic (SPC), where a ``sneaky generator'' deliberately produces flawed reasoning steps to challenge a ``critic'' model. This forces the critic to evolve its assessment capabilities, ultimately achieving performance that surpasses existing process reward models without manual step-level annotation. 
Zhou et al.~\cite{zhou2024reflect} proposed a two-player online game between a ``proposer'' that generates a response and a ``reflector'' that provides immediate, dense feedback by critiquing it. This dynamic interaction generates rich, supervisory signals that are absent in static feedback datasets. 
Xie et al.~\cite{xie2024learning} used a Stackelberg game framework to learn detoxification from non-parallel data. The insight of their work is the finding that the success of such methods is often highly dependent on the accuracy of the feedback signal, such as the toxicity classifier.

\subsubsection{Evolving the Reward Signal}

A static reward model (RM), no matter how well-trained, is inevitably exploitable. As the LLM policy improves, it can discover and exploit edge cases or loopholes in the RM's fixed reward function, a phenomenon known as reward hacking~\cite{skalse2022defining}. 
To counter this, researchers have reframed the interaction between the LLM and the RM as a dynamic game where the reward signal co-evolves with the policy. 
The Adversarial Preference Optimization (APO) framework~\cite{cheng2024adversarial} lets the LLM and RM update in a min-max game. The RM is trained to find outputs where the current LLM policy is poorly calibrated, and the LLM is then trained to improve on these adversarial examples. 
Its update also incorporates a KL divergence regularization term, ensuring it remains faithful to the original human preferences while adapting to the LLM's new distribution.
This dynamic can be formalized using frameworks from game theory, like bilevel optimization and Stackelberg games. STA-RLHF~\cite{STARLHF} models the interaction as a Stackelberg game where the LLM policy is the ``leader'' and the preference model is the ``follower.'' The policy makes the first move, and the preference model must best respond to it, forcing the policy to learn an alignment that is robust to an adaptive reward signal.
Shen et al.~\cite{shen2024principled} considered a more principled algorithmic framework, introducing a provably convergent first-order algorithm for such bilevel problems using a penalty-based method. 
Concurrently, Chakraborty et al.~\cite{chakraborty2024parl} developed a framework that, by precisely modeling the objective's dependency on policy-generated trajectories, improves the sample efficiency and mitigates reward over-optimization. 
These principled approaches are also being extended to more complex scenarios, such as Contextual Bilevel RL, which enables solving complex, real-world incentive alignment problems like tax design~\cite{thoma2024contextual}.

\subsubsection{Other Game-Theoretic Dynamics}

Game theory also provides novel frameworks for LLM alignment and generation that go beyond the data-reward dichotomy. 
One such area is cooperative and mixed-motive games. In the COEVOLVE framework~\cite{ma2024coevolving}, an LLM is fine-tuned by interacting with a copy of itself in a sequential cooperative game. 
Complementing this, Liao et al.~\cite{liao2024efficacy} provided empirical evidence that self-play is highly effective in non-zero-sum negotiation games. Beyond fine-tuning, evolutionary game theory is used to analyze emergent collective behaviors of LLM agent systems. 
Gemp et al.~\cite{gemp2024states} formulated the natural language dialogue generation as a game process. By applying equilibrium solvers, the method equips LLM with more stable and rational conversational strategies.

Another line of work applies game theory to the decoding process. The Consensus Game~\cite{jacob2024consensus} uses equilibrium search in a cooperative signaling game, where a Generator and Discriminator reconcile their predictions to find a consensus, leveraging no-regret learning to produce truthful and coherent outputs. 
The Peer Elicitation Games (PEG)~\cite{chen2025incentivizing} extends this idea by replacing a single agent discriminator with a multi-agent peer elicitation process.  
This equilibrium-seeking principle has been successfully applied to complex, embodied AI tasks; for instance, Yu et al.~\cite{yu2024vln} integrated it into a vision-language navigation system to reduce model hallucinations.
On the other hand, Chen et al.~\cite{chen2024decoding} proposed the Decoding Game, a theoretical framework that reimagines text generation as a zero-sum game against an adversarial ``Nature.'' Their analysis shows that this framing provides the first theoretical justification for the empirical success of heuristic methods like Top-k sampling. 
Zhang et al.~\cite{zhang2025strategic} reframed decoding as a Bayesian game between two internal LLM agents: a Generator and a Verifier. The two agents strategically interact to reach a ``Separating Equilibrium,'' a stable state where the verifier can reliably distinguish high-quality outputs from low-quality ones. This strategic decoding process acts as a powerful, training-free verification mechanism. 
Game dynamics can also steer generation at inference time, for instance by using solvers like Counterfactual Regret Minimization to guide dialogue toward less exploitable strategies~\cite{gemp2024steering}, or using Nash equilibrium concepts to dynamically control text attributes~\cite{sefeni2024game}.

Besides, game-theoretic mechanisms are being used to improve the efficiency of the alignment process. 
Zhang et al.~\cite{zhang2024vickreyfeedback} introduced an auction-based mechanism for collecting preference data, using principles from mechanism design to improve cost-efficiency.
Xie et al.~\cite{yi2025from} introduced Efficient Coordination via Nash Equilibrium (ECON), which recasts multi-LLM coordination as an incomplete-information game seeking a Bayesian Nash equilibrium. This framework allows each LLM to independently select responses based on its beliefs about co-agents, achieving a tighter regret bound and outperforming existing multi-LLM approaches. 

\begin{tcolorbox}[
    colback=white!5!white,      % 方框背景颜色 (这里是浅灰色)
    colframe=gray!75!black,     % 方框边框颜色 (这里是深灰色)
    title=\textbf{Discussions}, % 方框标题，加粗
    fonttitle=\bfseries,        % 标题字体加粗
    boxsep=5pt,                 % 方框内容与边框的距离
    arc=5pt,                    % 圆角半径
    outer arc=5pt,              % 外圆角半径
    left=5pt,right=5pt,top=5pt,bottom=5pt % 内边距
]
Although theoretical guarantees ensure desired properties, implementing these methods in practice poses significant challenges for robust performance. For example, computing the exact Shapley value is often intractable, and approximations suffer from high variance. Similarly, training self-play methods for value alignment or static problem-solving frequently encounters instability. Additionally, extracting preferences from highly heterogeneous datasets introduces further hurdles, such as incentive misalignment and moral constraints. Thus, substantial research opportunities remain to improve robustness, stability, and efficiency, building on the foundational ideas presented in this section.
\end{tcolorbox}

The game-theoretic methods used to enhance LLM interpretability, alignment, and adaptation do not exist in a vacuum. They are developed in response to the immense economic and social pressures that define the LLM ecosystem. The competition among stakeholders, the economics of data, and the deployment of LLMs as strategic actors in society create the demand for more robust, efficient, and aligned models. In Section 4, we will analyze this broader strategic landscape, using game models to characterize the multi-stakeholder competitions and societal impacts that motivate the technical advancements discussed herein.

%% file: sections_journal/tables/3_game_for_llm_improvement.tex
\begin{table}[t]
\centering
\caption{Application of Game-theoretic Methods for Improving Large Language Models.
% This table summarizes how key challenges in LLM development are addressed using specific frameworks from game theory, aligning with the structure of Section~\ref{sec:game_for_llm_algorithm}.
}
\label{tab:game_for_llm_improvement}

\resizebox{0.88\textwidth}{!}{
\begin{tabularx}{\textwidth}{@{} >{\raggedright\arraybackslash}p{3cm} >{\raggedright\arraybackslash}p{4cm} >{\raggedright\arraybackslash}X @{}}
\toprule
\textbf{Challenge} & \textbf{Core Game-theoretic Concepts} & \textbf{Game Formulation} \\
\midrule

\textbf{Model Interpretability}~\S\ref{sec:game_for_llm_algorithm-interpretability} & 
Cooperative Games \par (Shapley Value) & 
\textbf{Players:} LLM components (tokens, data points, layers, heads). \newline
\textbf{Goal:} Cooperate to generate the model's output. \\
\addlinespace

\rowcolor{lightgray}
\multicolumn{3}{@{}p{\dimexpr\textwidth-2\tabcolsep}}{%
    \textit{Spotlight:} Provides a principled method for credit assignment, enabling input attribution, data valuation, and model pruning. \newline
    \textit{Examples:} {TokenSHAP, Data Shapley, SV-NUP}~\cite{horovicz-goldshmidt2024tokenshap, ghorbani2019data, sun2025efficient}
} \\
\hline 
\addlinespace

\textbf{Aligning General Preferences}~\S\ref{sec:game_for_llm_algorithm-general} & 
Minimax Games \par (Nash Equilibrium) & 
\textbf{Players:} Two policies competing to be preferred by a human or learned oracle. \newline
\textbf{Goal:} Find a stable policy that cannot be consistently defeated (a Nash Equilibrium). \\
\addlinespace

\rowcolor{lightgray}
\multicolumn{3}{@{}p{\dimexpr\textwidth-2\tabcolsep}}{%
    \textit{Spotlight:} Overcomes limitations of scalar reward models, enabling robust alignment with complex (e.g., intransitive) preferences. \newline
    \textit{Examples:} {NLHF, SPO, DNO, MPO}~\cite{NLHF, SPO, DNO, wang2024magnetic}
} \\
\hline 
\addlinespace

\textbf{Capturing Preference Heterogeneity}~\S\ref{sec:game_for_llm_algorithm-heterogeneity} & 
Social Choice Theory \& \par Cooperative Bargaining & 
\textbf{Players:} User subgroups or annotators with diverse values. \newline
\textbf{Goal:} Aggregate preferences or negotiate outcomes according to formal fairness axioms. \\
\addlinespace

\rowcolor{lightgray}
\multicolumn{3}{@{}p{\dimexpr\textwidth-2\tabcolsep}}{%
    \textit{Spotlight:} Moves beyond monolithic alignment to design equitable systems that respect minority views and handle conflicting values. \newline
    \textit{Examples:} {MaxMin-RLHF, Axiomatic Analysis, Negotiative Alignment}~\cite{chakraborty2024maxmin, ge2024axioms, mushkani2025negotiative}
} \\
\hline 
\addlinespace

\textbf{Dynamic Adaptation}~\S\ref{sec:game_for_llm_algorithm-dynamic} & 
Competitive Self-Play, \par Stackelberg Games, \& \par Bilevel Optimization &
A co-evolving game with two primary forms, mapping to the section's structure:
\begin{itemize}[nosep, leftmargin=*, itemsep=0pt, topsep=0pt]
    \item \textbf{Evolving Data (\S\ref{sec:game_for_llm_algorithm-dynamic}.1):} A generator model creates challenging data for a critic or its past self.
    \item \textbf{Evolving Rewards (\S\ref{sec:game_for_llm_algorithm-dynamic}.2):} A "leader" policy optimizes against a "follower" reward model that finds its weaknesses.
\end{itemize} \\

\rowcolor{lightgray}
\multicolumn{3}{@{}p{\dimexpr\textwidth-2\tabcolsep}}{%
    \textit{Spotlight:} Replaces static components with a dynamic process that prevents reward hacking and enables continuous, autonomous improvement. \newline
    \textit{Examples:} {SPIN, STA-RLHF, Decoding Game}~\cite{SPIN, STARLHF, chen2024decoding}
} \\ 

\bottomrule
\end{tabularx}
}
\end{table}

%% file: sections_journal/4_game_models_for_llm.tex
\section{Characterizing LLM-related Events through Game Models}\label{sec:game_models}

As LLMs evolve from mere tools to active institutional players within markets, platforms, and information ecosystems, their behavior is no longer driven solely by technical objectives.
Game theory provides a rigorous framework for analyzing the increasingly complex strategic interactions surrounding the development, deployment, and societal integration of LLMs.
This section reviews recent advances in game-theoretic modeling applied to LLMs and categorizes them into two complementary streams:
Subsection~\ref{sec:game_models-development} addresses strategic games arising directly from the development and deployment lifecycle of LLMs, encompassing data acquisition, fine-tuning, platform economics, and content monetization.
In contrast, Subsection~\ref{sec:game_models-implication} examines broader societal impacts of LLMs, including platform-creator dynamics, policy-induced externalities, and economic coordination challenges.
The key problem settings, game notions, and summarized insights are presented in Table~\ref{tab:llm_game_models_summary}. 

\input{sections_journal/tables/4_game_model_for_llm}

\subsection{Multi-Stakeholder Competition and Cooperation in LLM Era}\label{sec:game_models-development}
The development and deployment of LLMs involve various stakeholders, ranging from data providers and annotators to model trainers, platform deployers, and end users. Strategic behavior is ubiquitous in this landscape, whether manipulation of preference reporting during model fine-tuning or intense competition among vendors for user attention and market share. 
Game-theoretic modeling offers a structured approach to exploring these behaviors, uncovering equilibrium structures, incentive misalignments, and design trade-offs.

\subsubsection{Strategic Preference Reporting in LLM Alignment}
Several studies examine strategic preference reporting in LLM alignment. Buening et al.~\cite{buening2025strategyproof} model RLHF as a principal-agent game, where the LLM developer (principal) relies on annotators (agents) to provide pairwise preference data for fine-tuning. Since annotators also derive utility from the resulting model, they may manipulate preferences to influence outcomes. Their analysis proves that no RLHF objective can simultaneously ensure strategyproofness and social welfare optimality.
Adopting a mechanism design perspective, Sun et al.~\cite{sun2024mechanism} similarly identify misreporting incentives under standard objectives and derive payment rules that enforce truthful reporting while maximizing welfare. In a simpler setting, Wu et al.~\cite{wu2025battling} formalize the Battling Influencers Game (BIG), showing it is a potential game where rational annotators exaggerate preferences in equilibrium to maximize influence. Together, these works reveal inherent incentive misalignment, necessitating carefully designed reward structures in alignment pipelines.
Liu et al.~\cite{liu2025humans, liu2025incentivizing} modeled scenarios where a principal cannot directly observe effort. They design and analyze contracts, such as bonus schemes using ``golden questions'' or linear/binary payment structures, to reward high-quality work. 
Both studies emphasize that without proper monitoring, even well-intentioned annotators may provide low-effort data, undermining alignment.

\subsubsection{Data Sharing and Model Release Strategies}
Game-theoretic approaches have emerged as powerful tools for analyzing strategic interactions in AI data-sharing, model development, and release strategies. 
In the domain of data sharing, Taitler et al.~\cite{taitler2025data} modeled the interaction between a content creation firm and a Generative AI (GenAI) platform as a Stackelberg game. The firm, acting as the leader, strategically controls information disclosure, while the AI platform, as the follower, determines how much data to acquire from external experts. Their equilibrium analysis shows that firms may be willing to pay GenAI platforms to use their data and identifies the conditions under which such agreements become Pareto-improving.
Focusing on model development, Laufer et al.~\cite{laufer2024fine} framed fine-tuning as a two-stage game between a generalist developer and a domain-specific specialist. This game involves sequential investment decisions followed by bargaining over revenue-sharing terms. Their findings demonstrate that a specialist's strategic decision to contribute, free-ride, or abstain hinges on the interplay of marginal returns and cost asymmetries.
To analyze competition among machine learning providers, Xu et al.~\cite{xu2025heterogeneous} introduced the Heterogeneous Data Game. This framework models providers that handle diverse data sources characterized by covariate and concept shifts. By identifying pure Nash equilibria, their work delineates the conditions that lead to distinct market structures: the non-existence of an equilibrium, convergence toward homogeneity, or specialization in heterogeneous niches.
Wu et al.~\cite{wu2025navigating} explored the strategic choice between open- and closed-source model releases, modeling it as a multi-agent game between open-source and proprietary developers. Their analysis reveals an ``innovation paradox'': while open-sourcing accelerates ecosystem-wide innovation, it can erode the competitive advantages of individual firms. The equilibrium outcome ultimately depends on factors like user demand elasticity and asymmetries in development costs.

\subsubsection{Pricing Strategies on Models and Outputs}
Economic modeling also extends to pricing strategies on models and outputs, frequently adopting a Stackelberg pricing framework. Here, LLM platforms play the role of leaders who pre-commit to pricing menus, while users, acting as followers, select options based on individual utility and task requirements.
Bergemann et al.~\cite{bergemann2025economics} studied the optimal pricing and product design for LLMs. Their economic framework considers variable operational costs for processing input and output tokens, the ability to fine-tune models, and diverse user needs and error sensitivities. They found that optimal pricing structures, often implemented through two-part tariffs, lead to higher markups for more intensive users, rationalizing observed industry practices like tiered pricing based on customization and usage levels.
Li et al.~\cite{li2025strategic} and Saig et al.~\cite{saig2024incentivizing} studied the pricing strategy for LLM users. 
The current pricing scheme, pay-per-token, is challenged by Saig et al.~\cite{saig2024incentivizing}, who found that companies may use a cheaper model rather than the model they claimed to use, which causes a moral hazard. To address such a problem, they introduce a pay-for-performance, contract-based framework that incentivizes quality. 
Li et al.~\cite{li2025strategic} modeled prompt pricing as a Stackelberg game between a platform and users with different prompt engineering skills. By incorporating ``prompt ambiguity,'' they derive an optimal pricing algorithm that adapts to user proficiency, improving platform payoff.
Mahmood~\cite{mahmood2024pricing} studied sequential price competition between two LLM-developing firms, setting prices for different tasks of model usage. The equilibrium analysis shows that the second mover can always achieve cost-effectiveness. Moreover, if the tasks are similar, the first mover may become cost-ineffective regardless of its pricing strategy.

\subsubsection{Advertising and Monetization within LLM}
Advertising and monetization within LLM interfaces represent a rapidly evolving area of research. Early explorations by Feizi et al.~\cite{feizi2023online} outline several potential advertising ecosystems, sparking further work on specific mechanisms. Duetting et al.~\cite{duetting2024mechanism} introduced a token-level auction where advertisers can bid to influence text generation, highlighting resultant challenges like the exposure problem. Building on this, Soumalias et al.~\cite{soumalias2024truthful} proposed MOSAIC, a truthful mechanism that aggregates advertiser preferences over entire outputs using Rochet payments and importance sampling.
Auction theory is also being adapted for Retrieval-Augmented Generation (RAG)~\cite{hajiaghayi2024ad, dubey2024auction}. In these models, advertisers hold text content and bid to have it integrated into the LLM's output. Recent studies are also exploring more complex dynamics; Banchio et al.~\cite{banchio2025ad} modeled dynamic auctions, revealing a temporal trade-off where delaying responses can create ``auction thickness'' to increase revenue. Meanwhile, Mordo et al.~\cite{mordo2024sponsored} investigated auctions that jointly compose sponsored and organic content to maximize social welfare rather than revenue alone. For a broader perspective, recent surveys~\cite{wu2024review} and position papers~\cite{wu2025advertising} map the entire design space for these novel systems.

\subsection{Framing the Societal Impact of LLMs}\label{sec:game_models-implication}

Beyond their immediate technical and economic applications, LLMs are profoundly reshaping broader societal structures. The widespread adoption of generative models introduces new dynamics of competition, cooperation, and strategic interaction across domains such as knowledge production, regulatory governance, platform design, and content labor markets. This subsection surveys game-theoretic research that models these macro-level effects, highlighting how emergent equilibria capture the unforeseen consequences of GenAI proliferation.

\subsubsection{Strategies of Autonomous LLM Agents}

A critical shift in modeling involves recognizing that GenAI systems can function as goal-directed, strategic agents. This perspective moves beyond viewing LLMs as passive tools, instead understanding them as economic actors whose implicit preferences, shaped by their training objectives, may misalign with human welfare. Immorlica et al.~\cite{immorlica2024generative} formalized this by treating GenAI as a consultant with a payoff function based on perceived helpfulness. They contend that even minor divergences between the AI's goals and user welfare can dramatically alter equilibria, leading to suboptimal outcomes such as overconfidence or persuasive bias, even when factual accuracy is maintained.

This strategic agency extends beyond theory. Taitler et al.~\cite{taitler2025selective} modeled ``selective response,'' where a GenAI purposefully withholds answers to niche queries to direct users toward human forums. This behavior generates fresh training data, establishing a beneficial data flywheel for the model, but it trades immediate user utility for long-term system optimization. Supporting these concerns, empirical work by Sabour et al.~\cite{sabour2025human} demonstrates that GenAI agents can manipulate human choices by strategically framing options, illustrating that algorithmic influence spans from recommendation to active persuasion. Collectively, these studies underscore the risks of deploying AI systems whose optimization criteria are not robustly aligned with human autonomy.

\subsubsection{Transformation in Data and Content Ecosystems}

The proliferation of GenAI has fundamentally altered incentives for human creators in data sharing and content production. Keinan et al.~\cite{keinan2025strategic} framed this challenge as a Prisoner's Dilemma, where individual creators must choose between sharing content for model training, risking competition from AI, or withholding it to preserve exclusivity. While cooperation yields higher platform-wide utility, each player's dominant strategy often involves defection, leading to suboptimal equilibria. This strategic tension can result in declining data quality unless carefully designed incentives, such as revenue sharing or content licensing schemes, are implemented.

Complementary research shows how these dynamics affect content diversity and creator viability. Gao et al.~\cite{gao2024pandora} demonstrated that while GenAI can increase average content quality, it also causes price erosion and reduces diversity due to output standardization, potentially displacing human creators from the market. The nature of this competition is further explored by Yao et al.~\cite{yao2024human}, who modeled human-AI interaction as a generalized Tullock contest for user attention. In this setting, players expend effort (\emph{e.g.,} producing engaging content) to win a probabilistic share of user attention, which translates into monetization or visibility. Their results indicate that stable coexistence is possible, with GenAI initially eliminating the least efficient human creators. Furthermore, the rapid evolution of GenAI imposes a significant adaptive burden on humans. Esmaeili et al.~\cite{esmaeili2024strategize} demonstrated that determining an optimal response strategy to a constantly evolving AI can be computationally intractable (NP-hard), formalizing the risk that human creators may be unable to adapt effectively in fast-moving content markets.

\subsubsection{System-Level Equilibria and Regulatory Challenges}

The strategic interactions among users, creators, and platforms can aggregate into unforeseen system-level equilibria and novel regulatory challenges. A stark example of emergent dysfunction is offered by Taitler et al.~\cite{taitler2025braess}, who adapted Braess's Paradox to GenAI deployment. In their model, a GenAI platform optimizing for revenue eventually erodes the quality of a human knowledge forum (\emph{e.g.,} Stack Overflow) by drawing away users. This degrades the quality of future training data, leading to a long-term outcome where all users are worse off, despite the GenAI's short-term utility.

Such emergent failures underscore the difficulty of effective governance. Laufer et al.~\cite{laufer2025backfiring} showed how well-intentioned but fragmented policies can backfire. For instance, imposing safety standards only on downstream AI fine-tuners may incentivize upstream developers to underinvest in safety, resulting in lower overall safety levels. This highlights the need for holistic regulation that aligns incentives across the entire development pipeline. Other systemic shifts include the ``flattening'' of supply chains, as modeled by Ali et al.~\cite{ali2025flattening}, where GenAI disintermediates gatekeepers, posing risks of surplus extraction and content homogenization. Finally, Xie et al.~\cite{xie2025algorithmic} examined the long-term dynamics between algorithmic decision-makers and humans, showing that misaligned incentives can lead agents to abandon self-improvement in favor of manipulation or exit, emphasizing the need for long-term alignment to achieve positive-sum outcomes.

\begin{tcolorbox}[
    colback=white!5!white,      % 方框背景颜色 (这里是浅灰色)
    colframe=gray!75!black,     % 方框边框颜色 (这里是深灰色)
    title=\textbf{Discussions}, % 方框标题，加粗
    fonttitle=\bfseries,        % 标题字体加粗
    boxsep=5pt,                 % 方框内容与边框的距离
    arc=5pt,                    % 圆角半径
    outer arc=5pt,              % 外圆角半径
    left=5pt,right=5pt,top=5pt,bottom=5pt % 内边距
]
Game-theoretic modeling provides a structured lens for analyzing strategic behaviors in the LLM ecosystem and understanding their broader societal implications. 
However, most existing results depend on simplified environments, typically involving a small number of rational agents with limited action spaces and clearly defined payoffs.
These abstractions can overlook emergent behaviors from large-scale interactions and fail to account for bounded rationality, institutional dynamics, and ambiguous motivations. 
Bridging the gap between theoretical predictions and empirical observations remains a challenging yet promising direction for future research.
\end{tcolorbox}

Thus far, we have explored how game theory is applied to evaluate, enhance, and characterize LLMs. 
As shown in our taxonomy, our survey considers a bi-directional relationship between the research on game theory and LLMs. 
In the next section, we will introduce studies that approach this intersection from a complementary perspective, discussing how they leverage LLMs' capabilities to advance classical game theory.

%% file: sections_journal/tables/4_game_model_for_llm.tex
\begin{table}[t]
\centering
\caption{Summary of Game Modeling for LLM-related Events}
\label{tab:llm_game_models_summary}
\small
\resizebox{0.80\textwidth}{!}{%  
\begin{tabular}{@{} >{\raggedright\arraybackslash}p{3.9cm} >{\raggedright\arraybackslash}p{2.9cm} >{\raggedright\arraybackslash}p{5.2cm} @{}}
\toprule
\textbf{Practical Scenarios} & \textbf{Game Frameworks} & \textbf{Key Findings/Insights}  \\
\midrule

\rowcolor{lightgray}
\multicolumn{3}{c}{Multi-Stakeholder Competition and Cooperation in LLM Era~\S\ref{sec:game_models-development}} \\
\addlinespace

Strategic Preference Reporting in LLM Alignment~\cite{buening2025strategyproof, wu2025battling, sun2024mechanism, liu2025humans, liu2025incentivizing} & Principal-Agent, Mechanism Design & Strategic misreporting harms alignment; trade-offs between strategyproofness and optimality; truthful reporting incentivized via payments. \\
\addlinespace
Data Sharing and Model Release Strategies~\cite{taitler2025data, laufer2024fine, xu2025heterogeneous, wu2025navigating} & Stackelberg Game, Repeated Game, Nash Equilibrium & Private and social incentives diverge in data-sharing; market structure depends on data heterogeneity and user behavior. \\
\addlinespace
Pricing Strategies on Models and Outputs~\cite{bergemann2025economics, mahmood2024pricing, li2025strategic, saig2024incentivizing} & Monopolistic Pricing, Stackelberg Game, Contract Theory & Pricing adapts to user types and skills; pay-per-token creates moral hazard, while pay-for-performance contracts align incentives for quality. \\
\addlinespace
Advertising and Monetization within LLM~\cite{feizi2023online, duetting2024mechanism, soumalias2024truthful, hajiaghayi2024ad, dubey2024auction, banchio2025ad, mordo2024sponsored, wu2024review, wu2025advertising} & Auction Theory, Mechanism Design & Truthful auctions at token or output level; dynamic auctions can increase revenue; mechanisms can co-optimize revenue and social welfare. \\
\midrule

\rowcolor{lightgray}
\multicolumn{3}{c}{Solving Intractable Game Problems with LLMs~\S\ref{sec:game_models-implication}} \\ 
\addlinespace

Strategies of Autonomous LLM Agent~\cite{taitler2025selective, immorlica2024generative, sabour2025human} & Economic Agent Models, Behavioral Game Theory & LLMs act as strategic agents with misaligned goals; they may strategically withhold info for long-term gain, leading to suboptimal outcomes. \\
\addlinespace
Transformation in Data and Content Ecosystems~\cite{keinan2025strategic, gao2024pandora, yao2024human, esmaeili2024strategize} & Prisoner's Dilemma, Contest Models, Computational Game Theory & Creators face a Prisoner's Dilemma in data sharing; GenAI competition erodes prices and diversity; human adaptation to AI is computationally difficult. \\
\addlinespace
System-Level Equilibria and Regulatory Challenges~\cite{taitler2025braess, laufer2025backfiring, ali2025flattening, xie2025algorithmic} & Network Games (Braess's Paradox), Stackelberg Games, Models of Regulation & Adding GenAI can degrade ecosystems (Braess's Paradox); fragmented regulation can backfire; long-term incentive misalignment leads to system failure. \\
\bottomrule
\end{tabular}
}
\end{table}

%% file: sections_journal/5_llm_for_game.tex
\section{Advancing Game Theory with Large Language Models}\label{sec:llm_for_game}

Traditional game theory strives for a comprehensive theoretical understanding within well-defined models. 
As a result, studies in this field often depend on formal, structured frameworks with limited strategy spaces and simplified communication protocols. 
While these constraints facilitate rigorous analysis and yield valuable theoretical insights, they also restrict the theory's ability to capture the complexities of real-world interactions.
Large language models, with their rich linguistic, reasoning, and representational capabilities, introduce a new computational paradigm. 
By using natural language as both an interface and a medium for strategic reasoning, LLMs enable researchers to revisit classic problems and explore domains that were previously intractable.
This emerging and promising area of research can be broadly categorized into two directions: the first (Subsection~\ref{sec:llm_for_game-expand}) explores how LLMs can expand traditional game-theoretic models, while the second (Subsection~\ref{sec:llm_for_game-solve}) focuses on leveraging LLMs to solve intractable game theory problems.

\subsection{Expanding Game Modeling with LLMs}\label{sec:llm_for_game-expand}

Traditional game-theoretic models often operate within rigid numerical and symbolic representations that fall short of capturing the nuance and fluidity of human interaction. 
By introducing natural language as a modeling medium, LLMs offer a transformative shift: \emph{enabling expressive, adaptive, and context-aware representations of strategic behavior}. 
This subsection explores how LLMs enrich the modeling landscape across three key domains: verbalized strategic interaction, preference elicitation in social choice, and semantic-enhanced economic mechanisms. 
Together, these works signal a move toward a more linguistically grounded and semantically rich game-theoretic paradigm.

\subsubsection{Verbalized Strategic Interactions}
Classical game theory's reliance on numerical information structures limits its ability to model human communication. Li et al.~\cite{li2025verbalized} addressed this by introducing a verbalized Bayesian persuasion (BP) framework for real-world games involving human dialogues. They represent BP as a mediator-augmented extensive-form game, with LLMs acting as sender and receiver. To solve this game efficiently, they develop a generalized equilibrium-finding algorithm that integrates LLMs with game solvers, incorporating verbalized commitment assumptions, obedience constraints, and information obfuscation. This approach enables game theory to model complex, language-driven interactions, enhancing its applicability to real-world scenarios like negotiations.

\subsubsection{Preference Elicitation in Social Choice}
In social choice, rigid preference elicitation with fixed alternatives restricts expressivity. Fish et al.~\cite{fish2024generative} proposed ``generative social choice,'' where LLMs generate textual options reflecting collective opinions, accommodating unforeseen alternatives. Their framework, supported by formal guarantees and empirical validation, demonstrates that LLMs can extract utilities from free-form text and create representative slates, achieving high agreement in applications like chatbot personalization. Boehmer et al.~\cite{boehmer2025generative} extended this with PROSE, a system generating diverse, cost-effective slates under budget and query accuracy constraints. Using public deliberation datasets (\emph{e.g.,} Polis), PROSE enhances scalability and user satisfaction in democratic processes, overcoming traditional social choice limitations.

\subsubsection{Semantic-Enhanced Economic Mechanisms}
Traditional auction and mechanism design models often assume fixed, black-box valuation functions, neglecting the semantic richness of real-world preferences. Sun et al.~\cite{sun2024large} introduced the Semantic-enhanced Personalized Valuation in Auction (SPVA) framework, where LLMs extract context-sensitive valuations from unstructured text (\emph{e.g.,} reviews, product descriptions) using LLMs. This reduces valuation noise and improves utility in Vickrey auctions. Similarly, Lu et al.~\cite{lu2024eliciting} extended mechanism design to natural language domains in peer-prediction settings, using LLMs to evaluate truthful reporting from free-form feedback, with applications in content moderation and community governance. Penna et al.~\cite{della2024natural} proposed Language Model Mechanisms (LMMs), where agents report in natural language, and LLMs compute outcomes and payments, maintaining incentive compatibility in high-dimensional, distributed environments. These advancements enrich economic mechanism design with semantic context.

\subsubsection{Emerging Perspectives}
These developments signal a paradigm shift in game-theoretic modeling. Lotfi et al.~\cite{lotfi2025rethinking} argued that LLMs' linguistic and adaptive capabilities challenge assumptions like fixed communication protocols and fully rational agents. In simulated negotiations, language-mediated interactions produce emergent behaviors, such as spontaneous coordination or implicit collusion, which standard models fail to predict. This shift toward semantically rich, language-based systems, where agents reason and renegotiate rules via natural language, expands the boundaries of game theory, enabling new insights into complex strategic interactions.

\subsection{Solving Intractable Game Problems with LLMs}\label{sec:llm_for_game-solve}

In addition to expanding how games are modeled, LLMs offer powerful tools for addressing computational bottlenecks that have long hindered the practical application of game theory. Their ability to generate, interpret, and formalize complex structures allows them to tackle problems traditionally deemed intractable. This subsection explores how LLMs contribute to four crucial areas: interpretable mechanism design, reducing cognitive load in allocation tasks, automated game modeling, and simulation-based reasoning for information disclosure. These contributions not only improve solution efficiency but also broaden the real-world applicability of game-theoretic analysis.

\subsubsection{Interpretable Mechanism Design}
Traditional mechanism design often produces opaque, black-box solutions. Liu et al.~\cite{liu2025interpretable} proposed a framework that casts mechanism design as a code generation task, using LLMs to produce human-readable pseudocode for heuristic mechanisms. This approach achieves competitive performance in complex design spaces, rediscovers classic mechanisms, and enhances interpretability, offering potential for discovering optimal mechanisms in intricate scenarios.

\subsubsection{Reducing Cognitive Burdens in Allocation Tasks}
High-dimensional preference reporting in allocation tasks, such as combinatorial auctions or course assignments, often overwhelms users, reducing efficiency. Soumalias et al.~\cite{soumalias2025llm} developed LLM-powered proxies that interpret one-shot natural language inputs to generate preference comparisons, lowering error rates and improving allocative efficiency. Similarly, Huang et al.~\cite{huang2025accelerated} showed that LLM-based proxies in combinatorial auctions, integrated with incremental revelation mechanisms, reduce query demands and achieve faster convergence compared to traditional learning-theoretic methods, streamlining complex allocation processes.

\subsubsection{Automated Game Modeling and Simulation}
Real-world strategic interactions described in natural language are challenging to formalize. Mensfelt et al.~\cite{mensfelt2024autoformalizing} introduced a framework for autoformalizing simultaneous-move games from textual descriptions, using one-shot prompting and syntactic feedback to create formal logic representations for analysis. Deng et al.~\cite{deng2025natural} extended this to imperfect-information games, employing a two-stage pipeline to identify information sets and construct extensive-form games, with a self-debugging module ensuring validity. Mensfelt et al.~\cite{mensfelt2024autoformalization} further enabled tournament-style simulations of strategies derived from natural language scenarios, completing a pipeline from informal input to executable strategic evaluation. These frameworks enhance game theory's adaptability to real-world scenarios like policy negotiations.

\subsubsection{Strategic Simulation and Information Disclosure}
LLMs enable novel simulation-based approaches to complex game-theoretic problems. Yin et al.~\cite{yin2025too} proposed InfoBid, a framework using LLM agents to evaluate how information disclosure policies affect auction outcomes. Their simulations reveal that classical assumptions, such as ``more information improves efficiency,'' may not hold with LLM bidders, as selective information sharing can lead to over- or underbidding. The limited ability of LLM bidders to model competitors' strategies highlights a gap between simulated and bounded rationality, underscoring the need to redesign strategies for LLM-mediated environments.

\begin{tcolorbox}[
    colback=white!5!white,      % 方框背景颜色 (这里是浅灰色)
    colframe=gray!75!black,     % 方框边框颜色 (这里是深灰色)
    title=\textbf{Discussions}, % 方框标题，加粗
    fonttitle=\bfseries,        % 标题字体加粗
    boxsep=5pt,                 % 方框内容与边框的距离
    arc=5pt,                    % 圆角半径
    outer arc=5pt,              % 外圆角半径
    left=5pt,right=5pt,top=5pt,bottom=5pt % 内边距
]
Most work in this section largely leverages the remarkable language generation capabilities of LLMs. 
However, LLMs are fundamentally prone to bias and hallucination, which can undermine the critical assumption of a stable and reliable ``oracle'' or ``solver'' in game-theoretic applications. 
This induces general imperfections and introduces unquantified and potentially systemic errors when LLMs act as simulators, preference elicitors, or mechanism designers, leading to real-world issues like unfairness in social choice or auction design. 
Under the case that an LLM might ``hallucinate'' game rules or reflect training data biases, it is challenging to make efficient verification and ensure the correctness and logical consistency of LLM-generated contents.
\end{tcolorbox}

%% file: sections_journal/6_future_directions.tex
\section{Challenges and Future Directions}\label{sec:future}
While existing research has made significant strides in the intersection between game theory and LLMs, several critical challenges remain unresolved. 
These limitations point to promising yet under-explored directions for advancing both theoretical frameworks and practical applications. 
In this section, we systematically identify these open problems and outline possible pathways for future research.

\subsection{LLM-based Agents with Comprehensive Game Abilities}
\textbf{Current Landscape:} Recent research has focused on evaluating and enhancing LLM agents' performance in specific, isolated game scenarios. 
For instance, studies have demonstrated significant improvements in strategic reasoning in matrix games~\cite{gandhi2023strategic}, Avalon~\cite{wang2023avalon}, bargaining~\cite{xia2024measuring}, and Werewolf~\cite{watanabe2024werewolf}.
Although some of the methods, such as strategic reflection or tool usage, are general in principle, their validation remains highly scenario-specific. 
Consequently, the performance improvement often fails to transfer effectively between different game genres or rule systems, limiting the development of truly generalist game-playing agents.

\textbf{Future Directions:} Based on this observation, a future direction is to develop LLM agents proficient in fundamental game-theoretic reasoning, capable of applying core principles across diverse game settings without requiring explicit customization for each new environment. 
Achieving this ambitious goal requires simultaneous advancements across multiple research fronts: (1) improved rule comprehension through better formal language understanding and symbolic reasoning integration, (2) more robust external environment modeling that can handle partial observability and stochastic transitions, and (3) sophisticated multi-agent reasoning frameworks that scale to varying numbers of participants with different behavioral patterns. 
As the goal is to build a generalist game-playing LLm, the validation data should be comprehensive, ranging from simple matrix games to complex imperfect-information games. 

\subsection{Moving Beyond Human-Oriented Evaluation Frameworks}
\textbf{Current Landscape:} 
One of the predominant approaches to evaluating the strategic capabilities of LLMs is through metrics designed to capture strategic optimality, such as Nash regret~\cite{brookins2024playing,mei2024turing,fontana2025nice} or $k$-level rationality~\cite{zhang2025k}. 
Interestingly, empirical observations reveal that LLMs often exhibit prosocial behaviors (Subsection~\ref{sec:game_for_llm_evaluation-observation}), frequently eschewing self-optimal strategies. 
One possible reason is that the RLHF has shaped LLM behavior toward more altruistic responses.
Moreover, the intrinsic training objective of LLMs, next-token prediction, diverges substantially from the principles underlying these evaluation metrics. As a result, it remains unclear whether success or failure on such metrics reliably reflects the true reasoning or strategic capacity of LLMs. This disconnect raises important questions about the adequacy and appropriateness of current evaluation paradigms.

\textbf{Future Directions:}
Developing evaluation frameworks specifically tailored to neural network-based agents is a valuable future direction. 
Merely repurposing benchmarks originally designed for humans is insufficient for capturing the unique behaviors and limitations of LLMs.
A meaningful starting point is the design of tasks that not only draw upon game-theoretic settings but also reflect the fundamental properties of LLM training. 
Consequently, evaluation metrics must also be custom-developed for LLMs, rather than borrowed wholesale from human cognitive testing.
An initial approach may involve the use of subjective or qualitative measures. 
And we should strive toward evaluation metrics that are robust, interpretable, and operationally meaningful. These attributes are crucial for ensuring that our assessments of LLM capabilities remain rigorous and actionable.

\subsection{Understanding LLMs' Strategic Behavior}
\textbf{Current Landscape:} 
Despite the improvement and evaluation of LLM agents in games, another valuable task is to provide a theoretical framework to characterize LLMs' behavior. 
For instance, Park et al.~\cite{park2025llm} provided insight into how LLM fails to act as a no-regret algorithm with the current supervised pre-training procedure. 
However, extending such theoretical characterizations to more complex strategic environments remains a significant open challenge. The vastness of strategy spaces and the combinatorial intricacies of game rules in realistic settings severely limit the feasibility of formal analysis.

\textbf{Future Directions:}
A robust theoretical understanding of LLMs in strategic contexts is critical for defining performance boundaries and guiding architectural design. A promising approach is to use abstraction and simplification to model essential game dynamics without their full analytical complexity, for instance, by analyzing critical subroutines or decision points in isolation. Moreover, tools from complexity theory, especially those related to the circuit complexity of Transformers, can be leveraged to establish formal limits on the strategic capabilities of these models. Ultimately, this line of inquiry could lead to more principled training methodologies that instill desired strategic behaviors.

\subsection{Capturing Cooperative Games in LLM Optimization}
\textbf{Current Landscape:} As discussed in Section~\ref{sec:game_for_llm_algorithm}, many studies applying game theory to LLM optimization have primarily focused on competitive game formulations. 
While competition offers a natural and tractable modeling approach, cooperation presents an equally promising avenue for advancing LLM capabilities, which is underexplored.

\textbf{Future Directions:} Incorporating cooperative game-theoretic principles into the training and optimization of LLMs could yield both novel theoretical insights and practical performance gains. For example, in Mixture of Experts (MoE) architectures, individual expert networks can be conceptualized as players in a cooperative game. 
Leveraging solution concepts such as the Shapley Value or the core could inform more principled router scheduling strategies, improving expert selection and load balancing while minimizing redundancy. 
Similarly, in ensemble learning and knowledge distillation settings, modeling sub-models as cooperative agents and applying fair credit assignment mechanisms could enhance collaboration among components, leading to improved generalization and computational efficiency.

\subsection{Modeling Cooperation Between Multi-LLMs and Humans}
\textbf{Current Landscape:} As reviewed in Section~\ref{sec:game_models}, prior research has predominantly focused on competitive or adversarial dynamics between LLMs and humans. These studies have shed light on important societal concerns, including persuasion, manipulation, and safety. Yet, cooperative interactions—particularly those involving formal game-theoretic modeling—remain significantly underexplored.

\textbf{Future Directions:} A promising and necessary research direction lies in understanding and designing cooperative frameworks involving multiple LLMs and human participants. Central challenges include constructing incentive-compatible mechanisms that encourage LLMs to coordinate effectively on human-assigned tasks while also accounting for their own modeled objectives. Developing a formal understanding of LLM agents' goals and behaviors is critical for bridging the gap between abstract theory and practical deployment. Progress in this area could lead to the design of AI systems that more robustly align with human values and intentions.

\subsection{Leveraging LLMs as Oracles to Extend Theoretical Game Models}
\textbf{Current Landscape:} As discussed in Section~\ref{sec:llm_for_game}, recent studies have demonstrated the potential of LLMs to extend classical game-theoretic models into more realistic domains involving natural language. The core insight is that the LLMs' sophisticated language understanding and generative abilities let them serve as oracles that instantiate otherwise abstract components of game models, functioning as certain rules.

\textbf{Future Directions:}  This approach paves the way for relaxing idealized assumptions by replacing theoretical constructs with practical, LLM-driven approximations. 
Such a substitution allows previously abstract models to be instantiated in real-world settings while preserving their approximate theoretical guarantees. 
Systematically exploring the use of LLMs as adaptable oracles could enable a new class of empirical, simulation-driven game-theoretic analyses.
A particularly impactful application lies in computational mechanism design—for instance, in simulating complex auctions where bidder preferences are subtle and difficult to formalize. LLMs can emulate realistic bidder behavior across varying mechanisms, supporting rapid iteration and refinement of designs that were previously analytically intractable.

%% file: sections_journal/7_conclusion.tex
\section{Conclusion}\label{sec:conclusion}
This survey has provided a comprehensive examination of the bidirectional interplay between game theory and large language models, addressing a significant gap in the existing literature. Through a novel structured taxonomy, we have systematically organized and elucidated the multifaceted relationship between these fields, highlighting their mutual reinforcement and synergistic potential. Our analysis demonstrates that game theory offers essential frameworks, such as equilibrium concepts, incentive design, and multi-agent interaction models, to formalize, analyze, and enhance LLM behaviors. Conversely, LLMs introduce unprecedented capabilities for simulating complex agents, approximating theoretical oracles, and scaling game-theoretic solutions in real-world environments. 
We hope this work inspires further exploration and collaboration at this dynamic intersection, paving the way for innovative developments in both domains.

\vspace{-0.5cm}